\documentclass[lettersize,journal]{IEEEtran}
\usepackage{amsmath,amsfonts}
\usepackage{algorithmic}
\usepackage{array}
\usepackage[caption=false,font=normalsize,labelfont=sf,textfont=sf]{subfig}
\usepackage{textcomp}
\usepackage{stfloats}
\usepackage{url}
\usepackage{verbatim}
\usepackage{graphicx}
\usepackage{multirow}
\usepackage{algorithm}
\def\BibTeX{{\rm B\kern-.05em{\sc i\kern-.025em b}\kern-.08em
    T\kern-.1667em\lower.7ex\hbox{E}\kern-.125emX}}
\usepackage{balance}
\usepackage{booktabs}

\newcommand{\etal}{\textit{et al.}}

\begin{document}
\title{Universal Multi-view Black-box Attack against Object Detectors via Layout Optimization}
\author{Donghua Wang, Wen Yao, Tingsong Jiang, Chao Li, Xiaoqian Chen \\
\thanks{Donghua Wang is with the College of Computer Science and Technology, Zhejiang University, Hangzhou, China. Wen Yao, Tingsong Jiang and Xiaoqian Chen is with the Defense Innovation Institute, Chinese Academy of Military Science and Intelligent Game and Decision Laboratory, Beijing, China. Chao Li is with the school of automation and electrical  engineering, Zhongyuan University of Technology, Zhengzhou, China. (Corresponding author: Wen Yao and Tingsong Jiang.)}
}

\maketitle

\begin{abstract}
Object detectors have demonstrated vulnerability to adversarial examples crafted by small perturbations that can deceive the object detector. Existing adversarial attacks mainly focus on white-box attacks and are merely valid at a specific viewpoint, while the universal multi-view black-box attack is less explored, limiting their generalization in practice. In this paper, we propose a novel universal multi-view black-box attack against object detectors, which optimizes a universal adversarial UV texture constructed by multiple image stickers for a 3D object via the designed layout optimization algorithm. Specifically, we treat the placement of image stickers on the UV texture as a circle-based layout optimization problem, whose objective is to find the optimal circle layout filled with image stickers so that it can deceive the object detector under the multi-view scenario. To ensure reasonable placement of image stickers, two constraints are elaborately devised. To optimize the layout, we adopt the random search algorithm enhanced by the devised important-aware selection strategy to find the most appropriate image sticker for each circle from the image sticker pools. Extensive experiments conducted on four common object detectors suggested that the detection performance decreases by a large magnitude of 74.29\% on average in multi-view scenarios. Additionally, a novel evaluation tool based on the photo-realistic simulator is designed to assess the texture-based attack fairly. 
\end{abstract}


\section{Introduction}
Deep learning, with the core of deep neural networks (DNNs), has profoundly impacted many research areas, such as computer vision, speech recognition, and natural language processes. However, DNNs have been demonstrated to be vulnerable to adversarial examples, which are crafted by small noises that are invisible to human observers but can deceive DNN models \cite{szegedy2014intriguing}. Therefore, adversarial examples impose potential security risk for the physically deployed DNN-based systems, especially in security-sensitive scenarios, such as autonomous driving and medical image diagnosing. An exhaustive exploration of such adversarial attacks-caused risk in advance is necessary to reduce the potential loss.

Currently, a line of studies on adversarial examples has emerged \cite{pgd2018towards,mim2018BoostingAA,li2022approximate,li2023adaptive,wang2023rfla}, and these works can be divided into white-box attacks and black-box attacks in terms of the adversarial's knowledge. White-box attacks assume the adversary can access full knowledge about the target model, such as architecture, training data, parameters, and so on. White-box attacks allow the adversary to leverage the model's gradient to optimize the adversarial example \cite{pgd2018towards,fgsm2015explaining,bim2016adversarial}, while it is impractical in the real world as the attacker can not access the deployed model's gradient. In contrast, black-box attacks prohibit the adversary from accessing the information about the model yet only allow them to query the model, which makes them match the physical scenario more practical. Nonetheless, current black-box attacks mainly focus on generating imperceptible \cite{li2022approximate,li2023adaptive} and image-specific adversarial perturbation \cite{wang2023rfla,tang2023natural} for each input image, whose flexibility is poor as it necessitates engender new perturbation for each input samples. Although some universal adversarial attack methods\cite{moosavi2017universal,mopuri2018generalizable,wang2023improving} can address the issue, these methods can not be physically deployed.

Physical deployable adversarial attacks are characteristic of the generated adversarial perturbation that can be manufactured and placed on the target cover object (e.g., pedestrian) in the physical world, which imposes practical security risks. Existing physical adversarial attacks can be roughly divided into the following two-fold in terms of the deployment of physical perturbation: sticker (patch)-based and camouflage-based physical attacks. The former performs physical attacks by sticking the sticker on a specific object area (e.g., the front side of the pedestrian \cite{thys2019fooling,huang2020universal}), but the attack will fail when the detector captures the image from the viewpoint where the adversarial sticker does not appear. In contrast, camouflage-based attacks \cite{wang2021dual,wang2022fca,duan2022learning,suryanto2022dta} generate the adversarial UV texture to modify the appearance of the 3D object by wrapping the texture over the surface of the 3D object, enabling them to maintain the attack effect on multi-view angles. However, most camouflage-based attacks belong to white-box attacks, limiting their usefulness. Although Wu \etal \cite{wu2020physical} proposed a black-box attack by leveraging the discrete search algorithm, which first optimizes the small image patch and then performs the enlarge-repeat process to construct the final adversarial texture. Nonetheless, the mosaic-like pattern of generated textures is conspicuous and easily attracts the attention of human observers.

Some approaches attempt to improve the inconspicuous of physical adversarial attacks, such as Liu \etal \cite{liu2019perceptual} exploited the generative model to generate the natural image, which is treated as the adversarial patch that is pasted on the sample to deceive the model. Doan \etal \cite{doan2022tnt} proposed a two-stage method to optimize the natural adversarial image patch against the classification model, in which the generator is trained to generate natural flower images at the first stage, and the input latent variable of the generator is optimized via the adversarial loss to ensure aggressive of the generated image.  Wei \etal \cite{wei2022adversarial} demonstrated that natural stickers can deceive the facial recognition system, where the paste location of stickers is found by their proposed search algorithm. Despite their success on the recognition task, adversarial attacks against the detection task proved more difficult than the recognition task \cite{chen2018shapeshifter}. Moreover, these methods are valid at specific viewpoints, limiting their effectiveness in the physical world.

\begin{figure}[t]
	\centering
	\begin{minipage}{.3\linewidth}
		\centering
		\includegraphics[width =1\linewidth]{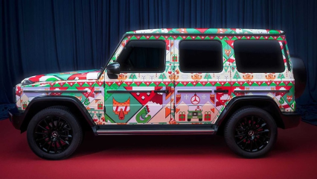}
	\end{minipage}
	\begin{minipage}{.3\linewidth}
		\centering
		\includegraphics[width =1\linewidth]{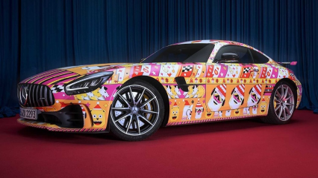}
	\end{minipage}
	\begin{minipage}{.3\linewidth}
		\centering
		\includegraphics[width =1\linewidth]{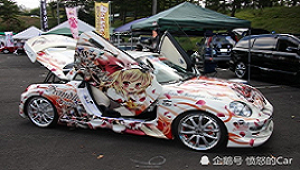}
	\end{minipage}
	\caption{Example of vehicle graffiti in the real world.}
	\label{fig:nautral_graffiti}
\end{figure}

Unlike optimizing the adversarial pattern of a single sticker for the object at the specific viewpoint, we optimize the layout of multiple sticker images in the UV texture of the object to perform universal adversarial attacks on multiple viewpoints. Note that the sticker can be arbitrary natural images, which renders the attack more inconspicuous as it looks like the vehicle with the graffiti, which is normal in the real world and can not attract human alert, as illustrated in Figure \ref{fig:nautral_graffiti}. In this paper, we propose a novel universal multi-view black-box attack on object detectors, which optimizes a universal adversarial UV texture for a 3D object via the designed layout optimization algorithm, rendering the generated image to deceive the object detector. More specifically, we treat the optimization problem of adversarial UV texture as a circle layout optimization problem, in which the rectangle is the container for image stickers randomly sampled from the preprepared image pool. The optimization objective is to find a universal adversarial UV texture wrapped over the 3D object to deceive the object detector from different viewpoints, and two layout constraint strategies are proposed to ensure the reasonable placement of image stickers. To obtain the optimal adversarial layout solution, we adopt the random search algorithm enhanced by the devised important-aware selection strategy. The optimization variable is the circle number, radius, and center, simplifying the constraint strategy. Additionally, we developed a novel evaluation tool based on the photo-realistic simulator (i.e., Unreal Engine) to assess the texture-based adversarial attack fairly. Finally, we conduct extensive experiments in both digital and simulated environments, showing the proposed method outperforms the comparison method significantly.

Our main contributions are listed as follows.
\begin{itemize}
\item We propose a novel universal multi-view black-box attack against object detectors via layout optimization, which engenders a universal adversarial UV texture that is valid in the multi-view scenario.
\item We formulate a circle layout optimization problem to obtain the optimal adversarial UV texture, which is solved by the random search algorithm enhanced by the designed important-aware selection strategy. Two layout constraints are designed to enable the reason for the placement of image stickers.
\item We devised a novel evaluation tool based on the photo-realistic simulator (i.e., Unreal Engine) to assess the attack performance of adversarial sticker attacks fairly.
\item Extensive experiments on both digital and simulated environments under different tasks show the effectiveness and extensibility of the proposed method. Moreover, we demonstrate that 3D objects with natural image graffiti can deceive object detectors in surround view. 
\end{itemize}
	
The rest of this paper is arranged as follows. The related works are briefly reviewed in Section \ref{sec:related_work}. Then, we formulate the problem statement in Section \ref{sec:problem}. The proposed method is exhaustly introduced in Section \ref{sec:method}. The evaluation tool is described in Section \ref{sec:tool}. Experiment and result analysis are discussed in Section \ref{sec:exp}. The conclusion is given in Section \ref{sec:conclusion}.

\section{Relate Works}
\label{sec:related_work}

In this section, we briefly review digital adversarial attacks and physical adversarial attacks.

\subsection{Digital Adversarial attacks}
Existing digital adversarial attacks can be categorized into white-box and black-box attacks in terms of whether the adversary can access the target model. The former assumes the adversary can access the full knowledge about the target model, which allows the adversary to exploit the gradient of the target model to develop adversarial attacks. Thus, the method used the gradient with respect to the model is regarded as the white-box attack. The most represented gradient-based approach is FGSM \cite{fgsm2015explaining}, which crafts the adversarial example along with the gradient ascend direction of task-specific loss with respect to the input in a single step. Since then, a line of variants of FGSM have emerged, such as multiple iterative steps (BIM) \cite{bim2016adversarial}, random initialization (PGD) \cite{pgd2018towards}, combined with momentum terms (MIM) \cite{mim2018BoostingAA} to improve the transferability, and input augmentation \cite{xie2021improving}. In contrast, black-box attacks prohibit the attacker from accessing the target model and only allow them to query the target model. Thus, a promising method to perform black-box attacks is the evolutionary algorithm, such as genetic algorithm \cite{alzantot2019genattack}, differential evolution (DE) \cite{lin2020black}, and particle swarm optimization (PSO) \cite{zhang2019attacking}. Despite obtaining certain success, their optimization involved amount of variables, because they were required to find the pixel-wise adversarial perturbation with the shape of the input image, which is time-consuming in practice. To reduce the optimization variables, Yang \textit{et al.} \cite{yang2020patchattack} optimized a small perturbation size and then interpolated it to the input shape, leading to a dramatic reduction in optimization variables. Rather than directly optimizing the pixel of perturbation, Andriushchenko \textit{et al.} \cite{andriushchenko2020square} utilized the random search algorithm to yield square-based perturbation, wherein the optimization variable is the position, side length, and fill color of the square. Li \textit{et al.} \cite{li2022approximate} proposed to exploit neighborhood samples around the input image to estimate the gradient, which is used to guide the update direction of adversarial examples. In their algorithm, the optimization variables solved by DE are the weights used to fusion the neighborhoods. Later, Li \textit{et al.} \cite{li2023adaptive} proposed to perturb partial pixels of input, which are determined by CAM \cite{zhou2016learning} that highlights the important pixel for model decision.

Another type of adversarial attack are universal attacks, which engender a single universal perturbation for the dataset. Moosavi \etal \cite{moosavi2017universal} first proposed an iterate update method to accumulate the perturbation for each image samples, where the final perturbation is called the universal adversarial perturbation. Concurrent, Mopuri \etal \cite{mopuri2017fast} devised a feature loss function to render the universal adversarial perturbation to disrupt the intermediate feature distributions. After that, a line of universal perturbation attack methods is proposed,  such as reducing the requirement of large-scale labeled training samples \cite{mopuri2018generalizable,mopuri2018ask}, improving the transferability across different models \cite{wang2023improving}.

Unlike optimizing the adversarial perturbation, some researchers attempt to search the sensitive location where the model is most vulnerable. Yang \textit{et al.} \cite{yang2020patchattack} extracted the class-specific textures by clustering Gram matrix of intermediate layers, which are used to paste on the image for realizing adversarial attack, in which the paste location is obtained by performing reinforcement learning. Dong \textit{et al.} \cite{dong2022viewfool} observed that natural objects under specific viewpoints in the real world are easily misrecognition by the model. To find such adversarial viewpoints, the author optimized the camera transformation (i.e., location and orientation) that made the model misclassification, then used the neural radiance field (NeRF) to generate realistic images based on that camera transformation. However, the above black-box attacks need to optimize the adversarial perturbation for each input, which is impractical. To generate universal adversarial perturbation, Ghosh \textit{et al.} \cite{ghosh2022black} proposed a novel black-box method to craft the universal adversarial perturbation against image classification tasks, but their method can not be physically deployed. The universal adversarial perturbation against the object detector under the black-box setting remains unexplored.

\subsection{Physical adversarial attacks}

The physical adversarial attacks can be divided into sticker (patch)-based or camouflage-based \cite{wang2022survey}. Sticker-based attack methods are designed to optimize an image sticker with the adversarial pattern, which will be manufactured (e.g., printed) and then deployed (e.g., hung \cite{thys2019fooling} or pasted \cite{tan2021legitimate,hu2021naturalistic,doan2022tnt}) on the object in the physical world. Thys \etal \cite{thys2019fooling} developed a patch-based attack against the pedestrian detector. Huang \textit{et al.} \cite{huang2020universal} proposed to optimize a universal adversarial patch with the initial natural image (i.e., dog), which is then printed out and pasted on the object repeatedly. Hu \etal \cite{hu2021naturalistic} used a generative model to engender a natural adversarial patch to attack the pedestrian detector. Wei \textit{et al.} \cite{wei2022adversarial} demonstrated the facial recognition model is susceptible to natural stickers. The sticker-based attacks show their limitation in multi-view angles. In contrast, camouflage-based attack methods were devised to optimize the UV texture of the 3D object directly, which is then wrapped over the surface of the 3D object, achieving better multi-view angle attack performance. Zhang \textit{et al.} \cite{zhang2019camou} devised a clone network to mimic the render-detection process as the physical render is not differentiable, then optimized the UV texture of the 3D object via white-box attacks. Wang \textit{et al.} \cite{wang2021dual} exploited the neural renderer and optimized the partial adversarial camouflage (i.e., vehicle's hood, roof, and side) by the devised attention suppress loss. Wang \textit{et al.} \cite{wang2022fca} took a step further than \cite{wang2021dual} and proposed a full coverage adversarial camouflage for the complex environment conditions (e.g., occlusion, multi-view) by maximizing the detector loss. Suryanto \etal \cite{suryanto2022dta} proposed an approximate renderer model to wrap the adversarial UV texture over the vehicle, which was optimized via gradient descent. Concurrent to \cite{suryanto2022dta}, Duan \etal \cite{duan2022learning} optimized the UV texture of the 3D object by maximizing the prediction score of the target class. Most approaches discussed above are white-box attacks that allow the adversary to utilize the target model's gradient to optimize the adversarial perturbation, while the black-box attacks are less explored. This work focuses on developing a universal multi-view black-box attack against object detection via layout optimization.

\section{Problem statement}
\label{sec:problem}
In this section, we first introduce the general problem statement of adversarial attack against object detection, and then we come up with the problem that this works to solve. 

\subsection{Adversarial attack against object detection}
Let $F$ indicate the trained object detection model to be attacked: $F: \mathcal{X} \rightarrow \mathcal{Y}$, where $\mathcal{X} \in \mathbb{R}^{C\times H \times W}$ ($C, H, W$ are the channel, height, and width of the image) and $\mathcal{Y} \in \mathbb{R}^d$, where $d$=6 for object detection, includes $x,y,w,h,F^{cls},F^{obj}$, the former four terms describes the predicted bounding box that include the coordinate of the center point, the width and height of the object, $F^{cls}$ and $F^{obj}$ are the predicted category and confident score of the object, respectively. In object detection, only the predicted $F^{obj}$ of the object larger than the assigned confident threshold $\tau$ is regarded as detected; otherwise, the object is regarded as undetected. We follow the previous work \cite{wang2021dual,wang2022fca} set the confident threshold $\tau$ to 0.5. The goal of adversarial attacks is to optimize a perturbation $\delta$ that decreases the $F^{obj}$ until beneath the predefined threshold $\tau$, while the perturbation does not arouse the attention of the human observer. Mathematically, the optimization of perturbation can be expressed as:

\begin{equation}
\min~~ \delta, \quad s.t. \quad F^{obj}(x+\delta) < \tau, \quad ||\delta||_p \leq \epsilon,
\label{eq:obj}
\end{equation}
where $||\cdot||_p$ indicates the $L_p$-norm, which confines $\delta$ to the $L_p$-norm sphere with radius of $\epsilon$ for ensuring the imperceptible of the adversarial examples.

\begin{figure}[t]
	\centering
	\begin{minipage}{.18\linewidth}
		\centering
		\includegraphics[width =1\linewidth]{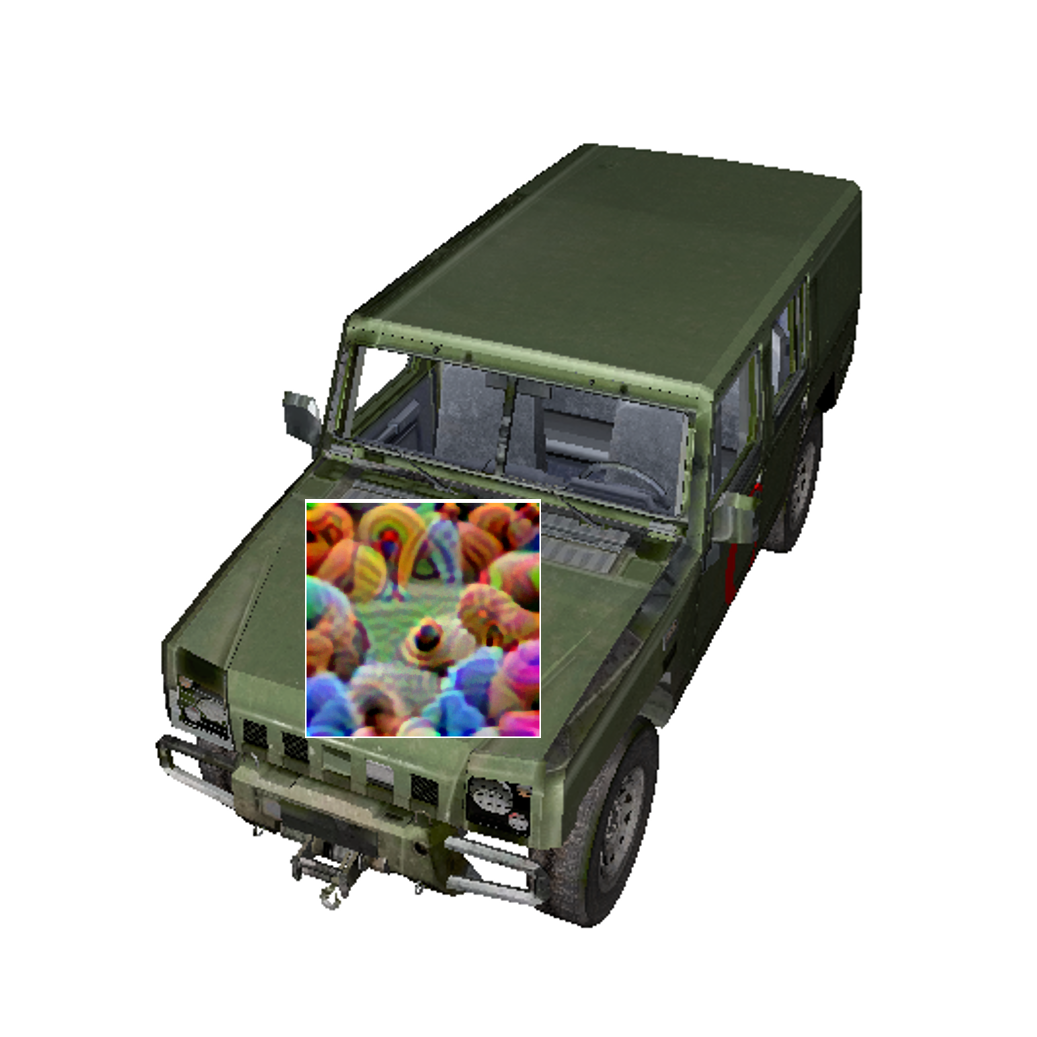}
		\centerline{\footnotesize (a)}
	\end{minipage}
	\begin{minipage}{.18\linewidth}
		\centering
		\includegraphics[width =1\linewidth]{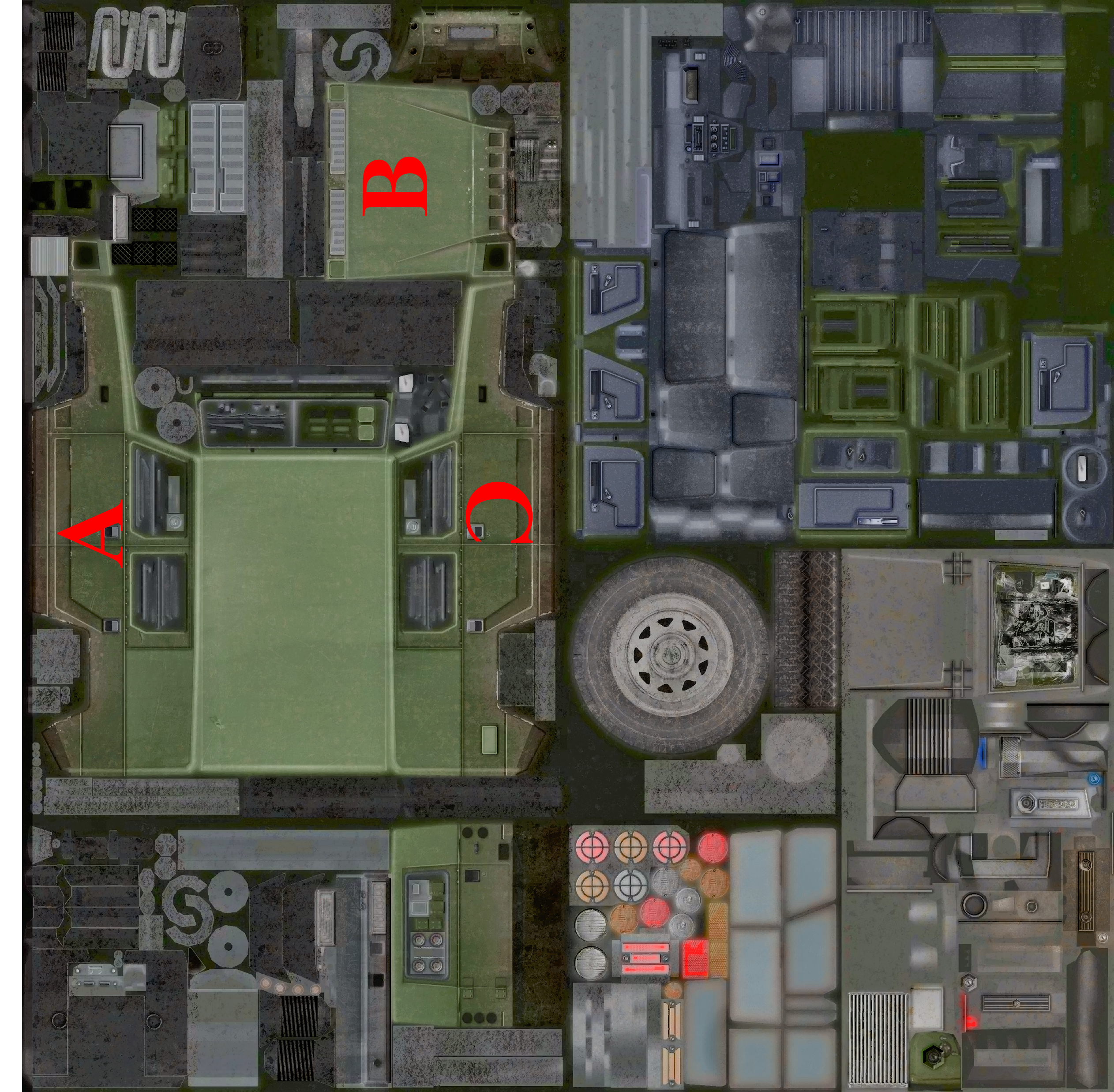}
		\centerline{\footnotesize (b)}
	\end{minipage}
	\begin{minipage}{.18\linewidth}
		\centering
		\includegraphics[width =1\linewidth]{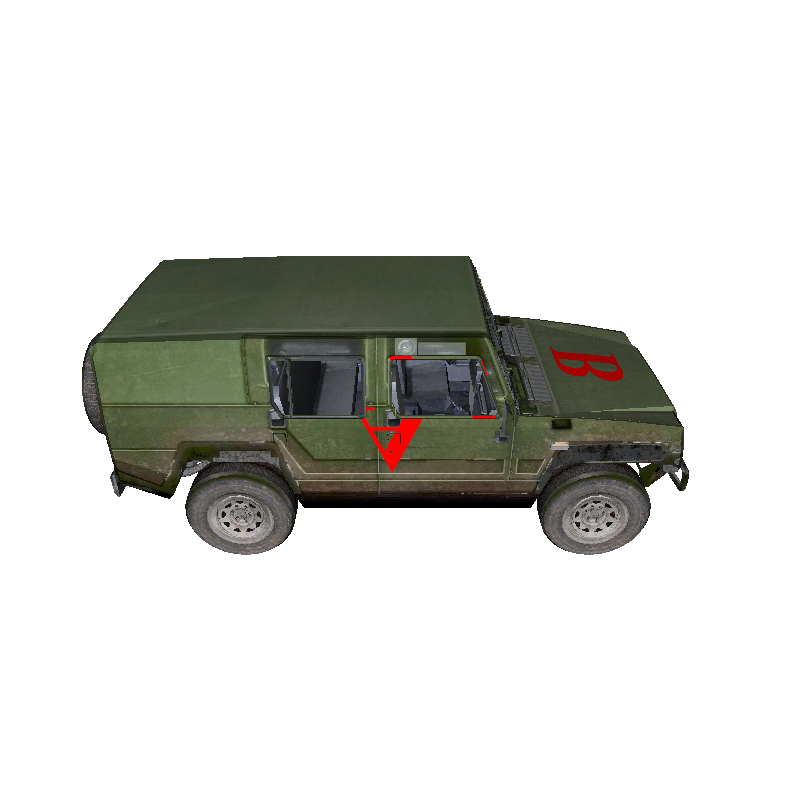}
		\centerline{\footnotesize (c)}
	\end{minipage}
	\begin{minipage}{.18\linewidth}
		\centering
		\includegraphics[width =1\linewidth]{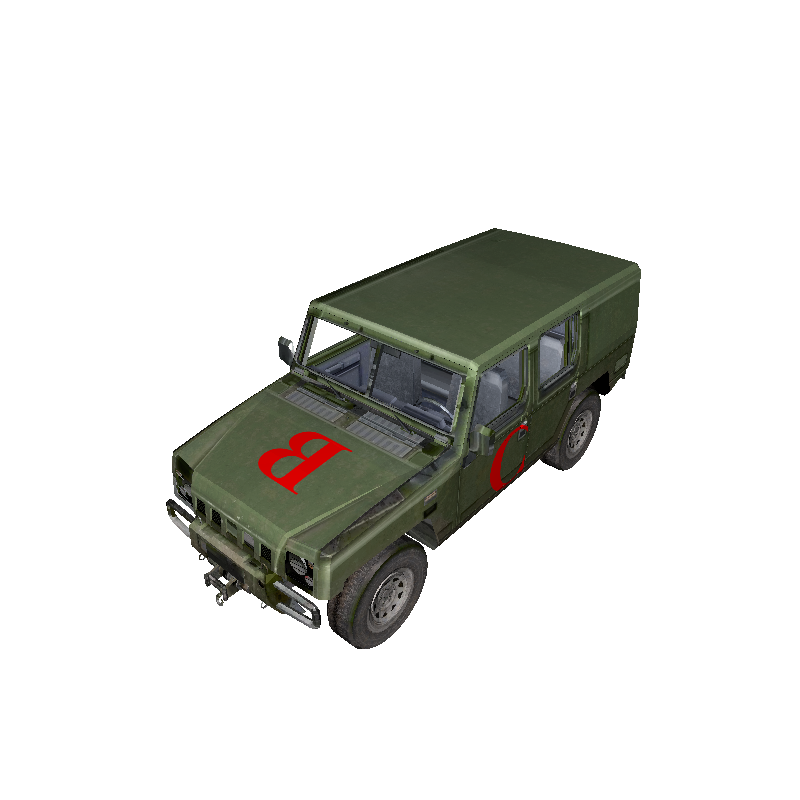}
		\centerline{\footnotesize (d)}
	\end{minipage}
	\begin{minipage}{.18\linewidth}
		\centering
		\includegraphics[width =1\linewidth]{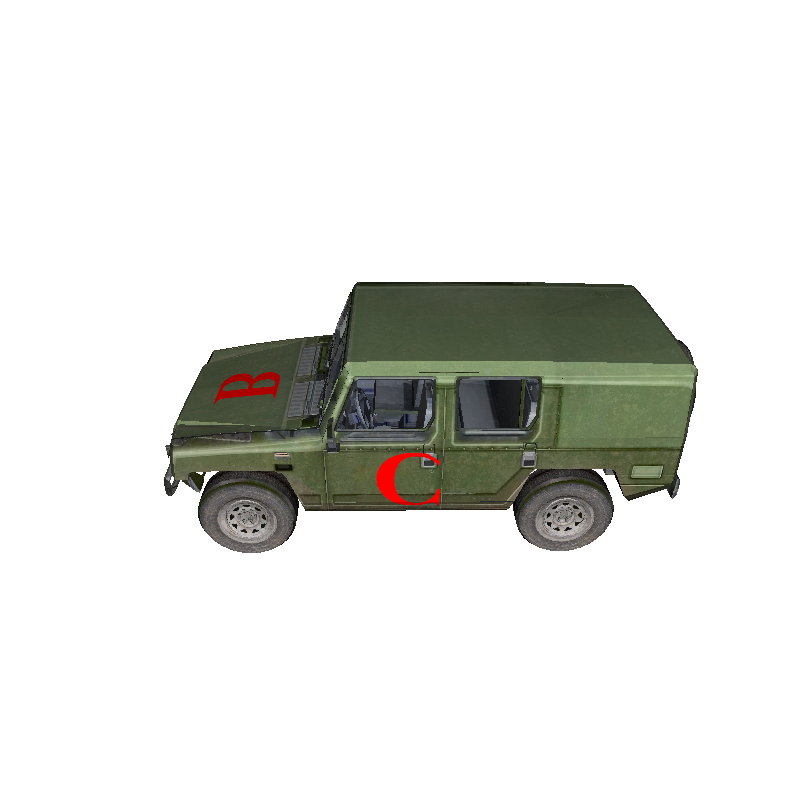}
		\centerline{\footnotesize (e)}
	\end{minipage}
	\caption{Illustration of sticker-based (a) and render-based (b-e) methods. From (b)-(e): (a) texture modified with the character; (b) right side position; (c) front position; (d) left side position. As we can see, only modifying a specific area (e.g., ``A") will fail to attack the rendered image (d).}
	\label{fig:renderer_example}
\end{figure}

\subsection{Universal multi-view attack}
The perturbation $\delta$ form defined in Equation \ref{eq:obj} is crucial for its physical deployment, such as when $\delta$ is the adversarial patch, it is unable to achieve multi-view attacks as they replace the partial pixel of the image with the adversarial patch (see Figure \ref{fig:renderer_example}(a)), ignore the non-plane characteristic of the 3D object. In contrast, we take non-plane characteristics into account and exploit the physical render to implement reasonable placement of image stickers. We wrap the adversarial UV texture constructed by multiple stickers over the 3D object via a physical renderer, changing the appearance of the 3D object. However, the 3D object has a different appearance observed from a different perspective, making a single adversarial sticker that can not work in multi-view conditions, as shown in Figure \ref{fig:renderer_example}(b)-(e). In this work, we aim to find a universal adversarial UV texture via the layout optimization algorithm, where the adversarial UV texture can be wrapped on the 3D object via the physical renderer and fool the object detector under a multi-view scenario.

Formally, given image stickers $\delta_{s}$, which are pasted on the UV texture $T$ of a 3D object with category $y_{gt}$ using an applier function $\mathcal{A}$, engender an adversarial UV texture $T_{adv}$ (i.e., $T_{adv}=\mathcal{A}(T, \delta_{s})$), which is then wrapped to the 3D object's mesh $M$ by a physical renderer $\mathcal{R}$, and output multiple adversarial examples $x_{adv}$ by using different camera transformations $\theta \sim \Theta$, $\theta$ contain $\mathtt{x, y, z; pitch, yall, roll}$, i.e., $x_{adv}=\mathcal{R}(M, T_{adv};\theta)$. Finally, the optimization object is to find appropriate image sticker $\delta_{s}$ to reduce the objectness score of the object detector for the specific 3D object under multi-view conditions, which can be expressed as

\begin{equation}
\underset{\delta_{s}}{\text{min}}~~\mathbb{E}_{\theta \sim \Theta} F^{obj}(\mathcal{R}(M, \mathcal{A}(T, \delta_{s});\theta); y_{gt}).
\end{equation}

\section{Methodology}
\label{sec:method}

In this section, we describe how we model the universal adversarial UV texture optimization problem as a circle-based layout optimization problem. Then, we elaborate on the proposed random search algorithm.

\begin{figure*}[t]
	\centering
	\begin{minipage}{.8\linewidth}
		\centering
		\includegraphics[width =1\linewidth]{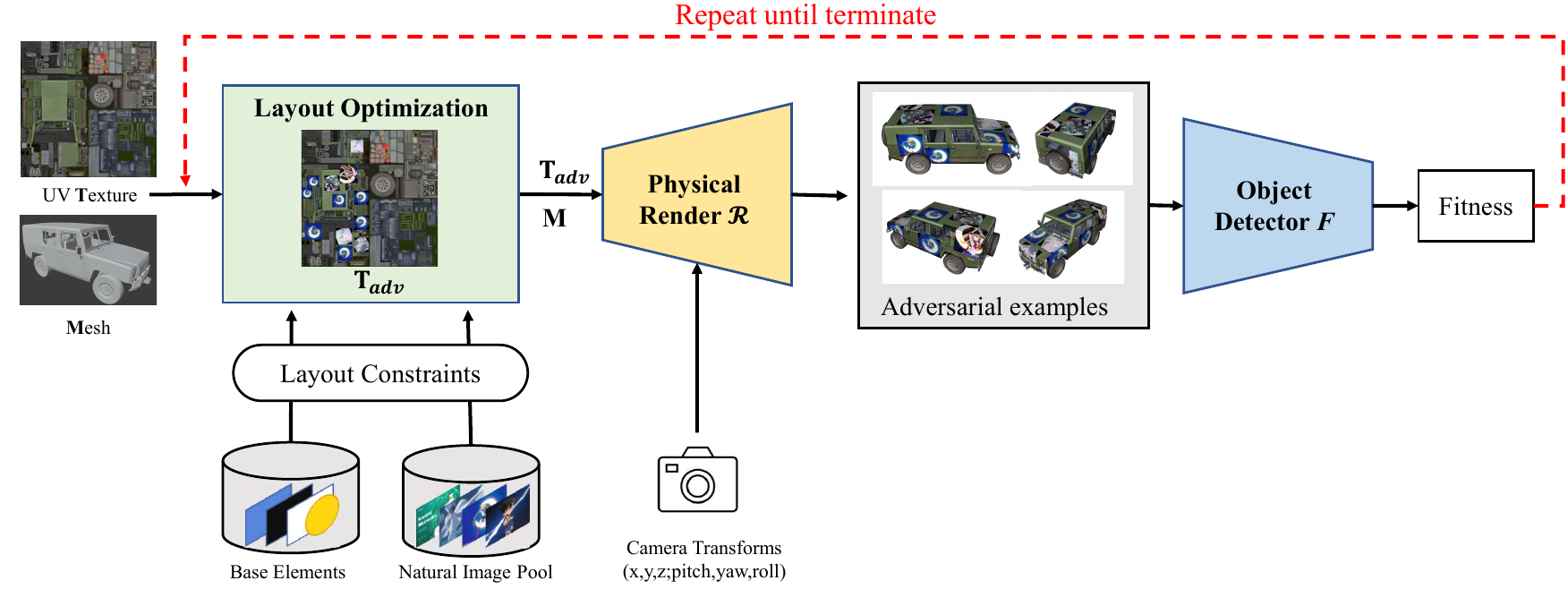}
	\end{minipage}
	\caption{Overview of the proposed method. The layout of adversarial patches in UV texture is optimized by choosing from the base elements (e.g., colored circle or square) or image sticker pool through layout constraints. With a physical renderer $\mathcal{R}$, the adversarial texture $T_{adv}$ is wrapped over the mesh $M$ of the 3D object and rendered to adversarial images $x_{adv}$, which are fed into the object detector $F$ for calculating fitness. }
	\label{fig:framework}
\end{figure*}

\subsection{Circle-based Layout Optimization}
Rather than optimize the pixel of adversarial stickers, we follow \cite{wei2022adversarial} to use the nature image as the adversarial sticker and optimize their number, size, and paste position on the UV texture, which can be regarded as the rectangle-based layout optimization problem. However, placing multiple rectangle stickers in UV texture inevitably causes overlaps, while the overlap control of the rectangle image is complexity. To address this issue, we formulate the problem as a circle-based layout optimization, which simplifies the overlap control during optimization. Moreover, the circle-based model can reduce the optimization variable, such as three variables (i.e., center position and radius) to determine the size and position of a sticker, but four variables (i.e., center position, width, and length) are required for the original rectangle-based model.

Based on the aforementioned discussion, adversarial stickers $\delta_{s}$ can be determined by the center position $p(p_x, p_y)$ and radius $r$. For multi-view attacks, we use $n$ stickers to modify the UV texture, ensuring each side of the vehicle's appearance is modified. We use $\mathcal{L}$ to indicate the sticker layout represented by $n$ sticker. Thus, $\mathcal{A}(T, \delta_{s})$ is reformulated as $\mathcal{A}(T, n, p, r)$, indicating that place $n$ stickers with radius $r$ at location $p$ on the UV texture $T$. Note that we aim to find the optimal sticker layout $\mathcal{L}$ to construct the universal adversarial UV texture $T_{adv}$, so adversarial stickers and $T_{adv}$ refer to the same thing, and we adopt $T_{adv}$ as following for simplicity. Finally, the layout optimization problem is expressed as follows.

\begin{equation}
\begin{split}
\rm find ~~& T_{adv} \\
\rm min  ~~& \mathbb{E}_{\theta \sim \Theta} F^{obj}(\mathcal{R}(M, \mathcal{A}(T, n, p, r);\theta); y_{gt}) \\
\rm s.t. ~~& n_{min} \le n \leq n_{max} \\
		~~& r_{min} \le r \leq r_{max} \\
     	~~& p \in \Omega \\
\end{split}
\label{eq:problem}
\end{equation}
where $n_{min}, n_{max}$ are the minimum and maximum allowable number of stickers in the UV texture, $r_{min}$ and $r_{max}$ are the lower and upper bound of the radius, which is determined by the area ratio of the sticker to UV texture; $\Omega$ refers to the region that allows for painting circles realized by constraints. Moreover, modificating the interior part of UV textures can not alter the 3D object's appearance (see Figure \ref{fig:constraints}(b)). For the selection of stickers, not only the natural image can be the sticker, but also the pure color block(namely base elements). Figure \ref{fig:framework} describes the pipeline of the proposed method. To address the layout optimization problem, we adopt the random search algorithm to find the optimal $T_{adv}$, which will be introduced as follows.

\subsection{Random Search for Universal Adversarial Texture}
\label{sec:random}
In this section, we elaborate on how we use the random search algorithm to find the universal adversarial texture for multi-view attacks represented by a circle-based layout optimization problem defined in Equation (\ref{eq:problem}), including layout initialization and layout optimization.

\subsubsection{Layout initialization}
We treat the construction of adversarial UV texture via adversarial stickers as the layout initialization, where the layout element is the sticker (e.g., sticker image or pure color image). More specifically, we first randomly sample the circle number $n$ from the $[n_{min}, n_{max}]$. For each circle, we obtain the circle radius $r$ determined by the predefined allowable modification area, followed by randomly sampling the circle's center $(x,y)$. Concretely, we bound the upper bound of the circle's area by exploiting the ratio $a$ of the circle's area to the UV texture's area $A$, such that we can get the circle radius by $\sqrt{A * a}$ where the $a$ should not exceed 10 percent. In the placement stage represented in $\mathcal{A}$, each circle should satisfy the overlap and mask constraint to make the placement coincidence with physical deployment, which will be discussed below. 

\begin{figure}[t]
	\centering
	\begin{minipage}{.3\linewidth}
		\centering
		\includegraphics[width =1\linewidth]{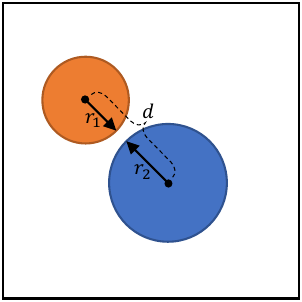}
		\centerline{\footnotesize (a) Overlap control}
	\end{minipage}
	\begin{minipage}{.6\linewidth}
		\centering
		\includegraphics[width =1\linewidth]{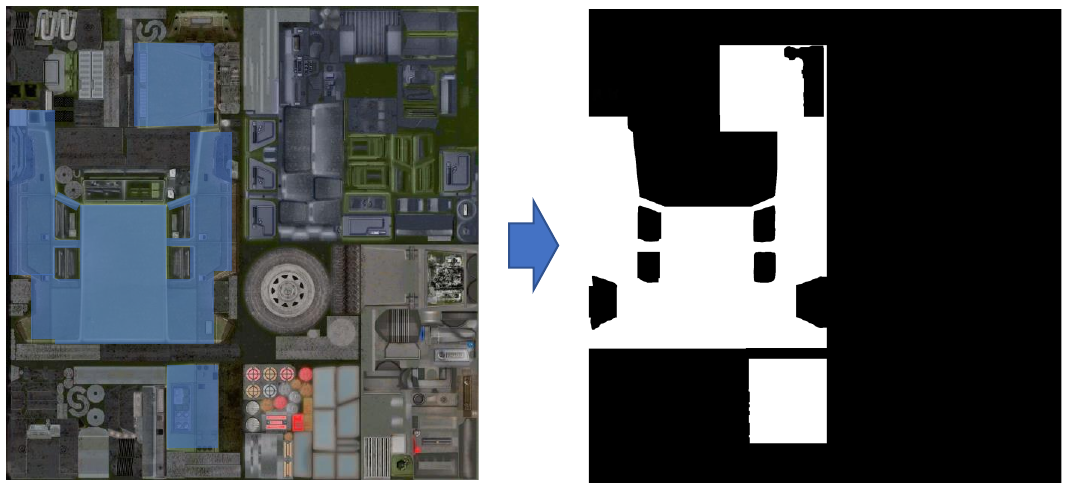}
		\centerline{\footnotesize (b) Mask constraint}
	\end{minipage}
	\caption{Layout constraints. From left to right: (a) Overlap control: avoid the overlap of the circle; (b) Mask constraint: confine the circle in the appearance region of the 3D object.}
	\label{fig:constraints}
\end{figure}

\textbf{Overlap constraint.} Overlap caused by multiple stickers can be easily solved by the circle-based layout. Specifically, we control the overlap by ensuring each spawn circle satisfies the following rule: the Euclidean distance of the circle's center between the new circle $(x_i, y_i)$ and all recorded circles $(x, y)$ is larger than the sum of their radius, which can be formulated as follows.

\begin{equation}
d = \sqrt{(x_i-x_j)^2 + (y_i-y_j)^2} \geq \gamma (r_i + r_j), \forall j \in \left\{ 1,..,|x| \right\},
\end{equation}
where $|x|$ denotes the number of recorded circles, and the $\gamma$ controls the overlap magnitude of two circles, which is set to 1. Figure \ref{fig:constraints} (a) provides the visualization explanation.

\textbf{Mask constraint.} Given the fact that some region in UV texture does not appear on the surface of the 3D object, such as vehicle interiors. To avoid the invalid search, we make a binary mask $\Omega$ to separate whether the region can appear on the appearance, where 1 in $\Omega$ denotes the region that allows being painted the circle (represented in white), while 0 does not (represented in black), Figure \ref{fig:constraints} (b) illustrates the mask constraint.

\begin{algorithm}[t]
	\caption{Layout checkout}
	\label{alg:layout_checkout}
	\textbf{Input}:current circle $l_{cur}$, recorded circle $\mathcal{L}$, texture mask $\Omega$, allowable overlap threshold $\gamma$ \\
	\textbf{Output}: Boolean
	\begin{algorithmic}[1] 
		\STATE $x_i,y_i,r_i \leftarrow l_{cur}$ 
		\STATE if $\Omega(x, y) = 0$
		\STATE \quad return false
		\FOR {$l ~{\rm in }~ \mathcal{L}$}
		\STATE $x_j, y_j, r_j \leftarrow l$ 
		\STATE $d_r = r_i + r_j$
		\STATE $d_c = \sqrt{(x_i-x_j)^2 + (y_i-y_j)^2}$
		\STATE if $d_c \leq d_r * \gamma$
		\STATE 	\quad return false
		\ENDFOR
		\STATE 	return true
	\end{algorithmic}
\end{algorithm}

\begin{algorithm}[t]
	\caption{Circle layout initialization}
	\label{alg:layout_initialization}
	\textbf{Input}: texture area $A$, texture mask $\Omega$, allowable number of patch $[n_{min}, n_{max}]$ and overlap threshold $\gamma$, are ratio $[a_{min}, a_{max}]$ \\
	\textbf{Output}: circle layout $\mathcal{L}$
	\begin{algorithmic}[1] 
		\STATE Randomly sample $n$ from $[n_{min}, n_{max}]$
		\STATE Initial circle layout $\mathcal{L}$
		\FOR {$itr=i,..., n$}
		\STATE Randomly sample ratio $a$ from $[a_{min}, a_{max}]$, where $a_{min}, a_{max}$ are the lower and upper bound of area ratio
		\STATE Calculate the radius $r$ by $\sqrt{A * a}$
		\STATE Randomly sample circle center coordinates $(x,y)$
		\STATE if Layout\_check(x, y, r, $\mathcal{L}$, $\Omega$, $\gamma$)  ~~~~$\triangleright$ Algorithm \ref{alg:layout_checkout}
		\STATE 	\quad $\mathcal{L} \leftarrow (x,y,r)$ 
		\STATE else
		\STATE \quad Resample
		\ENDFOR
	\end{algorithmic}
\end{algorithm}

\begin{figure}[t]
	\centering
	\begin{minipage}{.75\linewidth}
		\centering
		\includegraphics[width =1\linewidth]{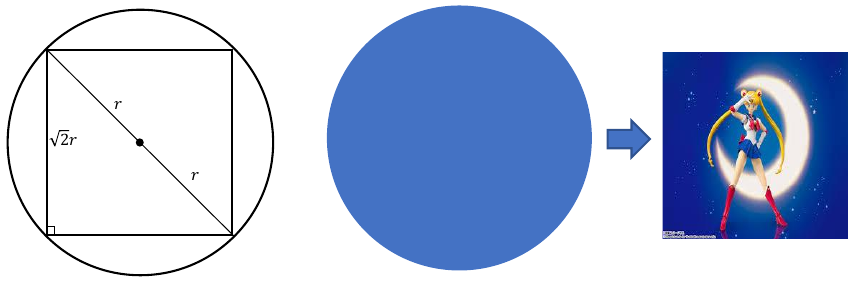}
	\end{minipage}
	\caption{Illustration of replacing the circle with other elements (e.g., rectangle and nature image) in layout optimization. From left to right: left: circle and the largest inscribed square; middle: pure circle layout; right: natural image layout.}
	\label{fig:natural_aware}
\end{figure}

To obtain the $n$ circle to construct the adversarial UV texture, we record the circle whose center is located in the mask $\Omega$ until sufficient circles are obtained. Meanwhile, every circle has to satisfy the overlap constraints. Algorithm \ref{alg:layout_checkout} describes the process of layout constraints. Once the circle layout $\mathcal{L}$ constructed, we replace each circle with the image to obtain the final adversarial UV texture. In this stage, we randomly select an image from the image pool for each circle. Then, we adopt the following strategy to place them on the UV texture: the size of the image is adjusted in terms of the size of the circle's largest inscribed square, where the side length is determined by $\sqrt{2}r$, as illustrated in Figure \ref{fig:natural_aware}.  Moreover, we adjust the image's orientation for a reasonable purpose. For example, the rendered image without adjusting the image sticker is odd (inverted image) in the real world (see Figure \ref{fig:renderer_example} (c)). Therefore, we designed a criterion to adjust the orientation of the sticker. Specifically, the sticker image rendered on the left side of the vehicle will be rotated 90 degrees clockwise, while the right side will be rotated 90 degrees counterclockwise. In summary, the whole process of layout initialization is described in Algorithm \ref{alg:layout_initialization}.

\begin{algorithm}[t]
	\caption{Adversarial UV texture optimization}
	\label{alg:random_search}
	\textbf{Input}: object detector $F$, physical renderer $\mathcal{R}$, camera transformation distribution $\Theta$, the 3D object of category $y_t$ comprise of mesh $M$ and texture $T$, mask constraint $\Omega$, maximum iterative step $Itr_{MAX}$ \\
	\textbf{Output}: adversarial texture $T^*_{adv}$
	
	\begin{algorithmic}[1] 
		\STATE $T_{adv} \leftarrow T$, $\mathcal{L}_{best} \leftarrow []$ 
		\STATE $f_{best} \leftarrow \sum_{\theta \in \Theta} \mathbb{I}(F_{cls}(\mathcal{R}(M,T;\theta)) = y_t)$
		\FOR {$itr=1,...,Itr_{MAX}$}
		\STATE  Obtain circle layout $\mathcal{L}$ with Algorithm \ref{alg:layout_initialization}
		\STATE  Update adversarial texture $T_{adv}$ with circle layout $\mathcal{L}$
		\STATE  calculate $f_{cur} \leftarrow  \sum_{\theta \in \Theta} \mathbb{I}(F^{cls}(\mathcal{R}(M,T_{adv};\theta)) = y_t)$
		\STATE  if $f_{cur} \leq f_{best}$
		\STATE  \quad $f_{best} \leftarrow f_{cur}$
		\STATE  \quad $T^*_{adv} \leftarrow T_{adv}$
		\STATE if $f_{best} = 0$
		 \STATE  \quad break
		\ENDFOR
	\end{algorithmic}
\end{algorithm}

\subsubsection{Layout Optimization}
Recall that our goal is to find a universal adversarial texture $T_{adv}$, which is wrapped over the 3D object and rendered into adversarial examples $x_{adv}$ via physical renderer $\mathcal{R}$ under different camera positions $\Theta$. At the same time, the objectness score of object detector $F$ on the generated adversarial examples beneath the threshold $\tau$. Therefore, we minimize the objectness score over rendered adversarial examples from different perspectives by optimizing the adversarial UV texture. Specifically, we devise the fitness score function $f$ as follows. 

\begin{equation}
	f = \sum_{\theta \in \Theta} \mathbb{I}(F^{obj}(\mathcal{R}(M, \mathcal{A}(T_{adv}, n, p, r);\theta);y_t) \le \tau ),
\label{eq:fitness_score}
\end{equation}
where $F^{obj}(\cdot)$ denotes the objectness score of the category $y_t$ the object detector $F$; $\mathbb{I}(\cdot)=1$ if true, and $\mathbb{I}(\cdot)=0$ if false. For the selection of $\Theta$, we sample the image per $2^\circ$ over $360^\circ$ and keep other settings fixed, collecting 176 images, which means that the universal adversarial UV texture $T_{adv}$ should maintain aggression on these images. 

There are two termination conditions: either achieving the max iteration number or the $T_{adv}$ can fool the object detector $F$ on all camera transformations $\Theta$. Algorithm \ref{alg:random_search} describes the optimization of the adversarial UV texture.

\subsection{Important-aware selection strategy}
In Section \ref{sec:random}, we adopt a randomly selected strategy to choose an image from the image pool, which may ignore the optimal image for the specific position, making it fails to obtain the best attack performance. To address this issue, we introduce an important-aware layout strategy. Specifically, we search for the best image from the image pool for each circle in the layout. With the important-aware selection strategy, we can find a better layout of the image sticker and obtain a higher attack performance. Mathematically, given the image sticker pool $S$, the best adversarial texture $T_{adv}$ can be obtained by solving the following problem.
\begin{equation}
G(s_i) = f(T_{adv}) - f(s_i \cup T_{adv}),~~s_i \in S
\end{equation}
where $G(s_i)$ is the gain function, we use $f(T_{adv})$ to denotes the fitness score of the adversarial texture $T_{adv}$, and $f(s_i \cup T_{adv})$ is the fitness score of adversarial texture $T_{adv}$ with the natural image $s_i$. By maximizing $G(s_i)$ for each layout position, we can obtain the optimal layout that could drop the fitness score the most. 

Moreover, before painting the image sticker on the texture, we introduce a random transform to improve the diversification by randomly rotating the image sticker within $[-30^\circ, 30^\circ]$. Algorithm \ref{alg:gen_texture} describes the generation method of the naturalistic-aware adversarial texture.

\begin{algorithm}[tb]
	\caption{Important-aware layout optimization}
	\label{alg:gen_texture}
	\textbf{Input}: circle layout $\mathcal{L}$, texture $T$, image pool $S$ \\
	\textbf{Output}: texture $T_{adv}$
	\begin{algorithmic}[1] 
		\STATE $T_{adv} \leftarrow T$ 
		\FOR  {$~l \quad in \quad \mathcal{L}$}
		\STATE $x, y, r \leftarrow l$ 
		\STATE $S_l \leftarrow G(s_l) = f(T_{adv}) - f(s_l \cup T_{adv}), ~~s_l \in S$
		\STATE if $p \leq 0.5$  ~~~~$\triangleright$ Random transformation
		\STATE \quad Rotate $s$ by randomly select angle from $[-30^\circ, 30^\circ]$
		\STATE Replace the circle with image $s$ at the coordinate $(x,y)$
		\ENDFOR
	\end{algorithmic}
\end{algorithm}

\section{Photo-realistic Evaluation Tool}
\label{sec:tool}
\begin{figure}[t]
	\centering
	\begin{minipage}{1.\linewidth}
		\centering
		\includegraphics[width =1\linewidth]{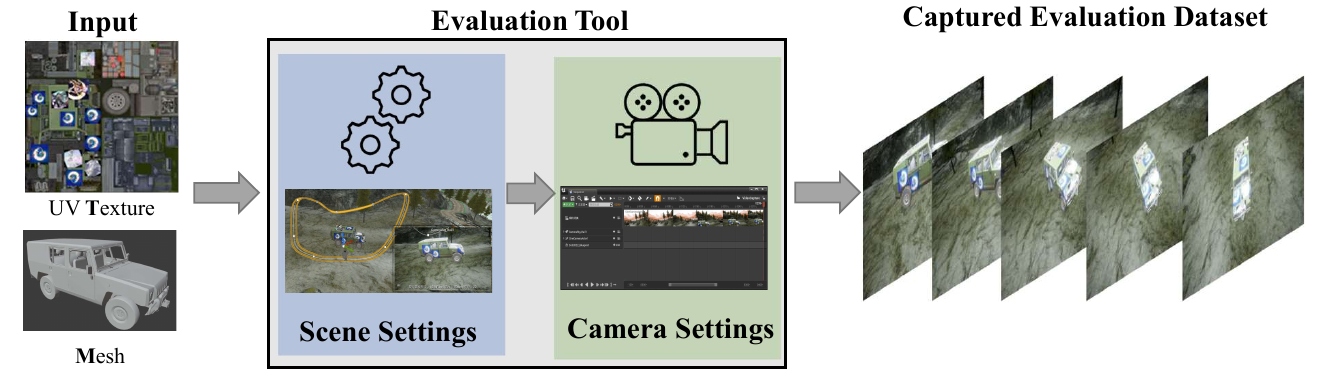}
	\end{minipage}
	\caption{The pipeline of the proposed evaluation tool. The tool requires the 3D object and the corresponding UV texture to generate multi-view simulated images.}
	\label{fig:tool}
\end{figure}

Currently, in physical adversarial attacks against object detectors, the lack of a uniform platform to assess the attack performance makes it hard to reproduce the existing approach due to the actual environmental discrepancy between the method and their follower who want to reproduce the physical attack. Moreover, the researcher usually evaluates physical attacks on the specific environment, while it is hard to reproduce by others. To address this issue, we presented a novel evaluation tool based on the photo-realistic simulator (i.e., Unreal Engine) to evaluate the attack performance of different attacks. Based on the simulator, we provide several evaluated backgrounds based on the scene (e.g., Landscape) provided by the simulator. For each background, we construct a camera rail around the target 3D object wrapped with adversarial texture, and then a camera is bound to the camera rail and towards the target object. Finally, a video sequence is recorded to construct the multi-view images, which are then used to assess the attack performance. The generated adversarial (UV) texture and the 3D object model are required to use the tool. With the proposed tool, anyone could reimplement other methods and make comparisons fairly. Figure \ref{fig:tool} describes the basic pipeline to use the evaluation tool. Therefore, we treat the simulated adversarial attack as the physical adversarial attack in this work.

\section{Experiments}
\label{sec:exp}
In this section, we first introduce the experiment settings. Then, we evaluate the proposed method on different object detectors under digital and physical conditions. We followed by conducting the ablation study on the influence of the image number of the seed image pool and the important-aware layout optimization. Finally, we extend our method to the warplane detection and traffic sign recognition (TSR) task.

\subsection{Settings}
\subsubsection{Target models} We evaluate the proposed method of vehicle detection and warplane detection on the following four object detectors: two one-stage (i.e., YOLOv5x \cite{glenn_jocher_2021_5563715}, and RetinaNet \cite{lin2017focal}) and two-stage detectors(i.e., Faster RCNN \cite{ren2015faster}, and Mask RCNN \cite{he2017mask}), provided by PyTorch \cite{paszke2019pytorch}. These detectors are trained on the MS COCO dataset \cite{lin2014microsoft}, which contains 80 categories. Moreover, We further evaluate the performance of the proposed method on two widely used TSR models: GTSRB CNN and LISA CNN. We follow \cite{liu2019perceptual} to resize the image resolution from $28 \times 28$ to $128\times128$ when training two models on GTSRB \cite{gtsrb2011} and LISA \cite{lisa2012} datasets, which obtain 95.06\% and 100\% accuracy on the test data, respectively. The reason for adjusting the resolution is that the naturalistic image will be severely distorted when resizing high-resolution images to low-resolution images, which makes it hard to achieve higher attack performance.

\subsubsection{Evaluation Dataset} We collect 176 images for the vehicle detection task with sampling per $2^\circ$ until covered $360^\circ$, where 160 images for the warplane detection task. Regarding the TSR task, we randomly select 100 naturally-looking ``STOP" sign images from the corresponding dataset.

\subsubsection{Evaluation Metric} To quantify the performance of the proposed method, we follow the prior works \cite{zhang2019camou,wu2020physical,wang2021dual} to adopt the P@0.5 as the measurement metric for vehicle detection and warplane detection. P@0.5 refers to the ratio of images correctly detected after attacks to the total test number when the confidence threshold is set to 0.5, which indicates the lower P@0.5, the stronger the attack. For TSR, we adopt the attack success rate (ASR) as the evaluation metric to represent the proposed method against the TSR model, indicating the higher the ASR, the better.

\subsubsection{Implementation Details} The physical renderer is the crucial component of the proposed method, which wraps the optimized adversarial UV texture over the 3D object. In this paper, we adopt the widely used physical renderer open3d library \cite{Zhou2018} as the physical renderer. Unless otherwise specified, we adopt the following default parameters for all experiments,  $n_{min}$ and $n_{max}$ are set to 5 and 15 for vehicle and warplane tasks, and 3 and 8 are for the traffic sign recognition task. $a$ is uniformly sampled from 0.001 and 0.1 for all tasks. The image numbers of the image pool are set to 30. The maximum iteration number $Itr_{MAX}$ set to 10000.

\subsection{Comparison result}
To investigate the effectiveness of the proposed method, we perform the attack by wrapping the adversarial texture stuck with the optimized image sticker layout on the vehicle. For comparison, we get inspiration from $RP_2$ \cite{eykholt2018robust} and SquareAttack \cite{andriushchenko2020square} and then modify them as the baseline method. $RP_2$ \cite{eykholt2018robust} stuck the white or black image patch on the 2D traffic sign image for deceiving TSR model. In contrast, SquareAttack \cite{andriushchenko2020square} is designed to optimize the invisible square adversarial perturbation by adding the pure color square block to fool the image classifier. Although these works achieve certain success in the original task, they are hardly applied to our tasks involving 2D to 3D transformation. Therefore, we combine the proposed method with their elements (e.g., square and white-block patch) to optimize the adversarial texture, and we treat these two methods as baseline approaches. Specifically, in our work, we replace the circle area with their image patch to mimic their methods, called BlackWhite \cite{eykholt2018robust} and SquareAttack \cite{andriushchenko2020square}. We also compare the proposed method with existing adversarial attacks against object detectors, which include CAMOU \cite{zhang2019camou}, UPC \cite{huang2020universal}, ER \cite{wu2020physical}, DAS \cite{wang2021dual}, FCA \cite{wang2022fca}, CAC \cite{duan2022learning}, and DTA \cite{suryanto2022dta}. Note that these methods are white-box attacks, so we only use their generated adversarial pattern or camouflage to construct the adversarial texture to perform black-box attacks for comparison. We treat the pure circle element as our baseline, and we select the adversarial texture generated by the proposed method using natural images (i.e., AdvPatch, Animal, and Cartoon) with the best attack performance for evaluation. For the adversarial texture constructed by natural images, we utilize the adversarial patch generated by the existing approaches, which may still be aggressive based on adversarial perturbation with transferability. In addition, we also adopt two public available image categories containing cartoon images and animal images. With a physical renderer, we wrap the optimal adversarial texture over the vehicle and render it under different views as adversarial examples to be evaluated. 

The experiment results are listed in Table \ref{tab:comparison_result}. As we can observe, on the one hand, the proposed method (i.e., Circle) outperforms the baseline method BlackWhite and SquareAttack by 16.76\% and 4.54\% in terms of the average degradation of P@0.5. At the same time, the pure color elements are unable to obtain better attack performance. On the other hand, AdvPatch outperforms all comparison methods, achieving a maximum decrease of 74.29\% in terms of the average P@0.5 over four detectors. Besides AdvPatch, although Cartoon falls behind the CAC, the discrepancy is small (i.e., 6.25\%). The reason may be attributed to CAC being designed to generate the full coverage adversarial texture under the white-box attack setting. However, the adversarial texture generated by Cartoon is more stealthiness than CAC.  Moreover, we observe that approaches devised to generate full coverage adversarial texture (i.e., FCA and CAC) exhibit better transferability than other methods (except for AdvPatch), particularly for attacking two-stage detectors (i.e., Faster RCNN and Mask RCNN), where the average degradation caused by FCA and CAC  are 55.12\% and 54.26\% in terms of P@0.5, surpass other methods in a large margin. Such observation indicates the advantage of the full coverage adversarial attack method in multiview conditions. Additionally, we find an interesting phenomenon that some adversarial textures would boost the detection performance of the detector, such as ER boosting the detect performance by 0.57\% for Faster RCNN and DAS and DTA enhancing the detect performance by 5.68\% for RetinaNet.

Regarding the image-based adversarial texture, the AdvPatch obtains the best attack performance, achieving the maximum degradation on P@0.5 on RetinaNet by 88.07\%. The possible reason is that AdvPatch naturally contains adversarially due to the transferability characteristic of adversarial perturbation, despite not being trained on the used target detector. Although Animal performs better than the baseline methods on average, it's falling behind the AdvPatch and Cartoon, which indicates that the content of the images has also impacted the attack performance. On the other hand, we observe the robustness discrepancy of different detectors varying. One-stage detectors display weaker than two-stage detectors under the proposed attack. Specifically, the average P@0.5 degradation of YOLOv5x and RetinaNet are 57.01\% and 60.8\%, while Faster RCNN and Mask RCNN are 40.72\% and 21.88\%, respectively.

\begin{figure}[t]
	\centering
	\begin{minipage}{.8\linewidth}
		\centering
		\includegraphics[width =1\linewidth]{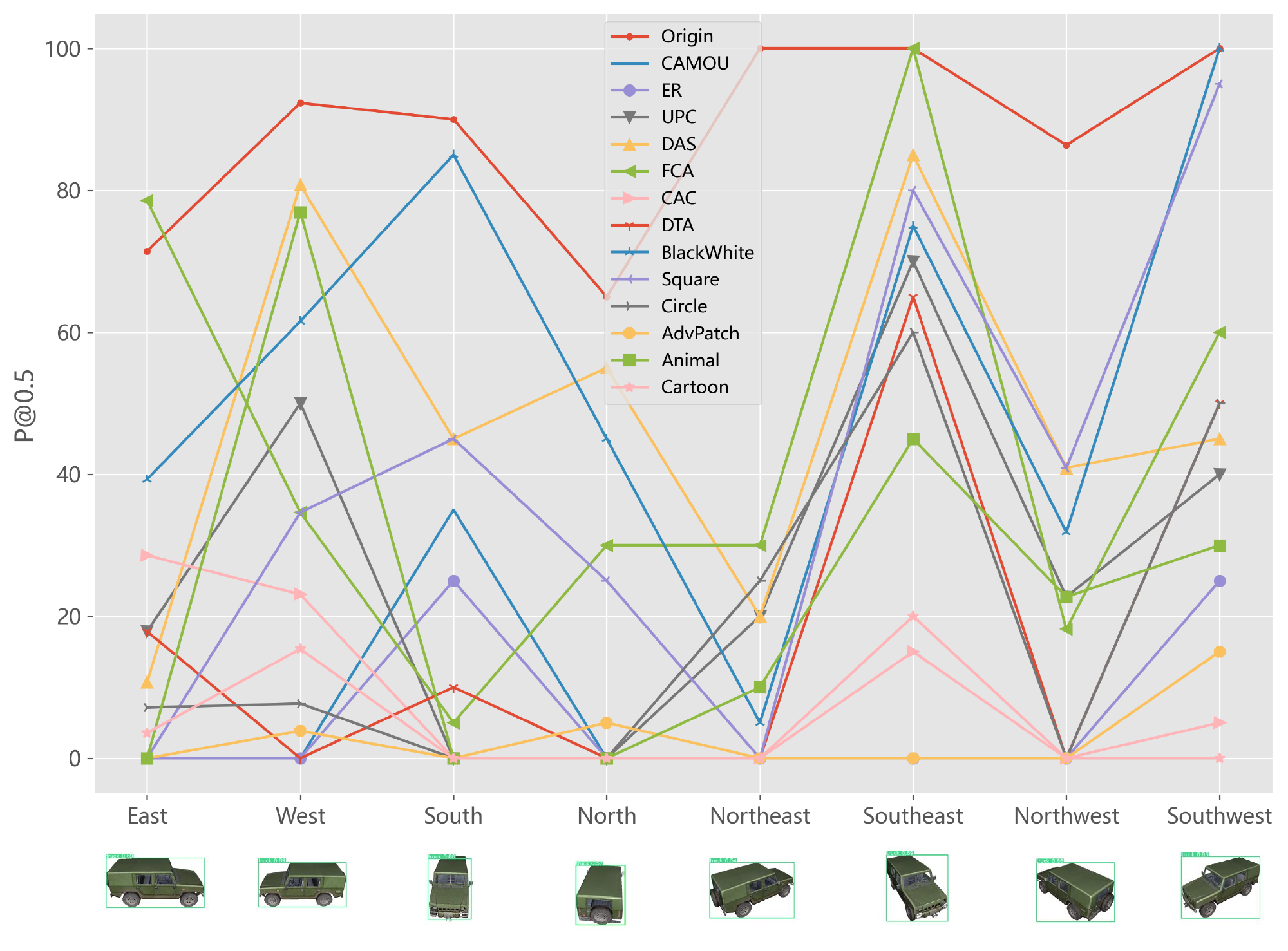}
	\end{minipage}
	\caption{Attack performance changes with direction. Zoom in for details.}
	\label{fig:asrvsangles}
\end{figure}

To analyze the influence of rendered viewpoints on attack performance, we regroup the rendered image into eight groups in terms of the rendered vehicle's heading, then statistics the change of P@0.5. Figure \ref{fig:asrvsangles} illustrates the evaluation results. We can observe that adversarial examples in the west, southeast, and southwest groups exhibit worse attack performance, where the P@0.5 are 29.88\%, 47.31\%, and 39.62\%, respectively. In contrast, the P@0.5 of the other five groups is less than 20\%. One possible reason is that the images in the west, southeast, and southwest display the strong vehicle contour characteristic. By contrast, the adversarial examples in the northeast group display the best attack performance, resulting in 91.54\% degradation of detection performance. 

\begin{table}[t]
\centering
\scriptsize 
\setlength\tabcolsep{5pt}
\caption{Comparison results of various \textit{digital attacks} in terms of P@0.5 (\%). \textbf{Bold} item highlights the best result, where the item in the bracket denotes the gain of P@0.5.}
\label{tab:comparison_result}
\begin{tabular}{ccccc}
\hline
          & YOLOv5x 				   & Faster RCNN			     & Mask RCNN 			   	  & RetinaNet \\ \hline
Raw       & 87.50 					   & 99.43				         & 100.0      				  & 92.05     \\
BlackWhite \cite{eykholt2018robust}      & 61.93($\downarrow$25.57)  & 75.57($\downarrow$23.29)   & 89.77($\downarrow$10.23)    & 55.11($\downarrow$38.64)     \\
SquareAttack \cite{andriushchenko2020square}  & 41.48($\downarrow$46.02)  & 52.84($\downarrow$46.02)   & 87.5($\downarrow$12.5)    & 51.7($\downarrow$42.05)     \\ \hline

CAMOU \cite{zhang2019camou}   & ~3.98($\downarrow$83.52)   & 97.73($\downarrow$~1.70)    & 93.18($\downarrow$6.82)    & 86.36($\downarrow$~7.39)     \\
UPC \cite{huang2020universal} & 27.84($\downarrow$59.66)   & 80.68($\downarrow$18.75)    & 90.34($\downarrow$9.66)    & 60.23($\downarrow$33.52)     \\
ER\cite{wu2020physical}       & ~5.68($\downarrow$81.82)   & 100.0($\uparrow$0.57)       & 100.0($\downarrow$~0.00)   & 15.35($\downarrow$17.05)      \\
DAS \cite{wang2021dual}       & 17.05($\downarrow$70.45)   & 98.30($\downarrow$~1.13)    & 97.16($\downarrow$2.84)    & 99.43($\uparrow$~5.68)         \\
FCA  \cite{wang2022fca}       & 45.45($\downarrow$42.05)   & 54.84($\downarrow$46.59)    & \textbf{36.36}($\downarrow$63.64)   & 22.73($\downarrow$71.02)         \\
CAC \cite{duan2022learning}   & 10.23($\downarrow$77.27)   & 44.89($\downarrow$54.54)    & 46.02($\downarrow$53.98)   & 26.14($\downarrow$67.61)     \\
DTA \cite{suryanto2022dta}    & 17.05($\downarrow$70.45)   & 98.30($\downarrow$~1.13)    & 97.16($\downarrow$2.84)    & 99.43($\uparrow$5.68)     \\ \hline
Circle         & 26.7($\downarrow$60.8)  & 57.95($\downarrow$40.91)     & 88.64($\downarrow$11.36)    & 42.05($\downarrow$51.7)     \\
AdvPatch  & \textbf{~2.84}($\downarrow$84.66)  & \textbf{34.09}($\downarrow$65.34)& 40.91($\downarrow$59.09)  & \textbf{5.68}($\downarrow$88.07)         \\
Animal    & 23.86($\downarrow$63.64)  & 68.18($\downarrow$32.25)     & 77.84($\downarrow$22.16)   & 21.02($\downarrow$72.73)     \\
Cartoon   & ~5.11($\downarrow$82.39)  & 56.82($\downarrow$42.61)     & 68.18($\downarrow$31.82)   & 22.16($\downarrow$71.59)    \\ \hline
\end{tabular}
\end{table}

\begin{figure*}[t]
	\centering
	\begin{minipage}{1.\linewidth}
		\centering
		\includegraphics[width =1\linewidth]{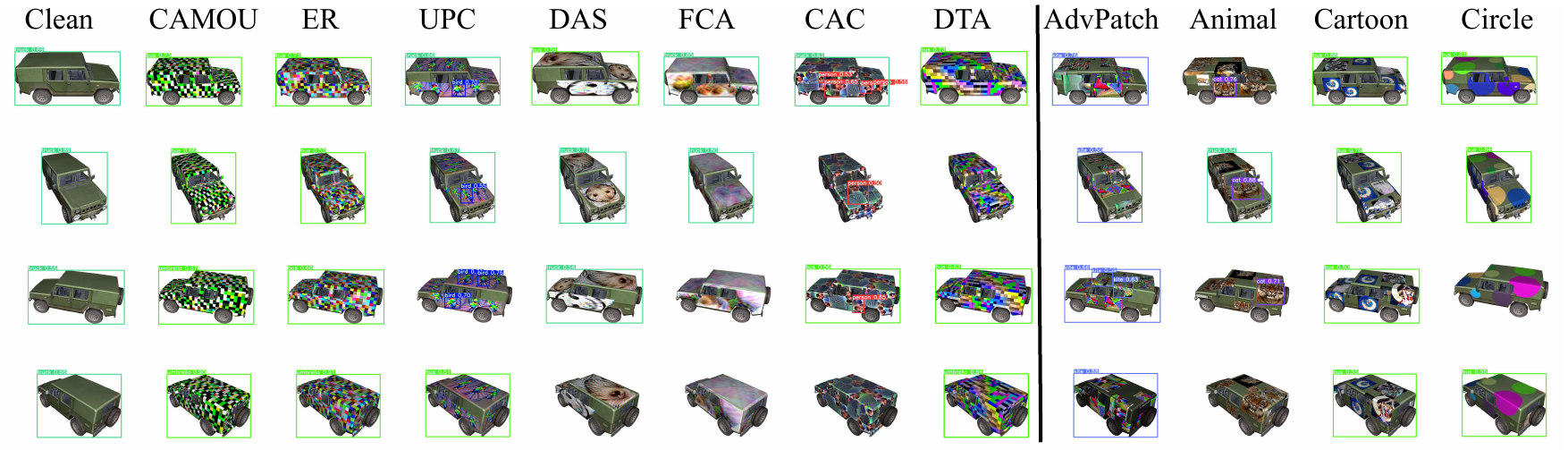}
	\end{minipage}
	\caption{Visualization of various adversarial textured vehicles (``truck"). The detection results are given by YOLOv5x. Zoom in for details.}
	\label{fig:visual_examples}
\end{figure*}

We also provide some adversarial examples rendered by different viewpoints in Figure \ref{fig:visual_examples}. As we can see, the adversarial examples generated by the proposed method can fool the detector into making the wrong detection in most cases (11/12). In contrast, comparison methods fail to deceive the detector. More importantly, compared with the mosaic-like adversarial texture generated by comparison methods \cite{zhang2019camou,wu2020physical}, the proposed method displays a better naturalness and stealthiness, which increases the risk of the proposed method when applied to the real world as people tend to ignore the security risk of scrawl painting, and result in significant accidents.

\begin{table}[]
\centering
\scriptsize
\setlength\tabcolsep{5pt}
\caption{Comparison results of \textit{simulated attacks} in terms of P@0.5 (\%). \textbf{Bold} item highlights the best result for comparison methods and proposed method respectively, where the item in the bracket denotes the derogation of P@0.5.}
\label{tab:simluated_attack}
\begin{tabular}{ccccc}
\hline
               & YOLOv5 & Faster RCNN & Mask RCNN & RetinaNet \\ \hline
RAW            & 76.33  & 97.00       & 100.0    & 87.33     \\
BlackWhite \cite{eykholt2018robust}    & 59.00($\downarrow$17.33)      & 86.33($\downarrow$10.67)       & 90.33($\downarrow$~9.67)    & 81.33($\downarrow$~6.00)     \\
SquareAttack \cite{andriushchenko2020square}  & 32.00($\downarrow$44.33)      & 83.00($\downarrow$14.00)       & 91.33($\downarrow$~8.67)    & 74.00($\downarrow$13.33)     \\ \hline
CAMOU \cite{zhang2019camou}       & 25.33($\downarrow$51.00)      & 83.67($\downarrow$13.33)    & 80.67($\downarrow$19.33)   & 64.33($\downarrow$23.00)     \\
UPC \cite{huang2020universal}   & 35.33($\downarrow$41.00)      & 84.67($\downarrow$12.33)    & 91.00($\downarrow$~9.00)   & 68.67($\downarrow$18.66)     \\
ER\cite{wu2020physical}          & 19.33($\downarrow$57.00)      & 85.67($\downarrow$11.33)    & 76.67($\downarrow$23.33)   & 71.33($\downarrow$16.00)      \\
DAS \cite{wang2021dual}          & 60.33($\downarrow$16.00)      & 89.33($\downarrow$7.67)     & 94.33($\downarrow$5.67)    & 66.67($\downarrow$20.66)         \\
FCA  \cite{wang2022fca}          & 57.33($\downarrow$19.00)      & 82.00($\downarrow$15)       & 87.00($\downarrow$13.00)   & 45.00($\downarrow$42.33)         \\
CAC \cite{duan2022learning}      & 23.67($\downarrow$52.66)      & \textbf{38.33}($\downarrow$58.67) & \textbf{49.00}($\downarrow$51.00) & \textbf{14.00}($\downarrow$73.33)   \\
DTA \cite{suryanto2022dta}        & \textbf{16.33}($\downarrow$60.00) & 80.67($\downarrow$16.33)   & 77.67($\downarrow$22.33)    & 76.00($\downarrow$11.33)     \\ \hline
Circle         & \textbf{29.00}($\downarrow$47.33) & 75.67($\downarrow$21.33)   & 83.67($\downarrow$16.33)    & 63.33($\downarrow$24.00)     \\
AdvPatch       & 34.67($\downarrow$41.66)          & \textbf{74.00}($\downarrow$23.00)   &\textbf{81.00}($\downarrow$19.00)  	   & \textbf{47.67}($\downarrow$39.66)     \\
Animals        & 68.00($\downarrow$~8.33)          & 89.33($\downarrow$~7.67)     & 94.33($\downarrow$5.67)    & 64.00($\downarrow$23.33)      \\ 
Cartoon        & 34.67($\downarrow$41.66)   	   & 81.00($\downarrow$16.00)     & 86.67($\downarrow$13.33)   & 66.33($\downarrow$21.00)     \\ \hline
\end{tabular}
\end{table}

\begin{figure*}[t]
	\centering
	\begin{minipage}{1.\linewidth}
		\centering
		\includegraphics[width =1\linewidth]{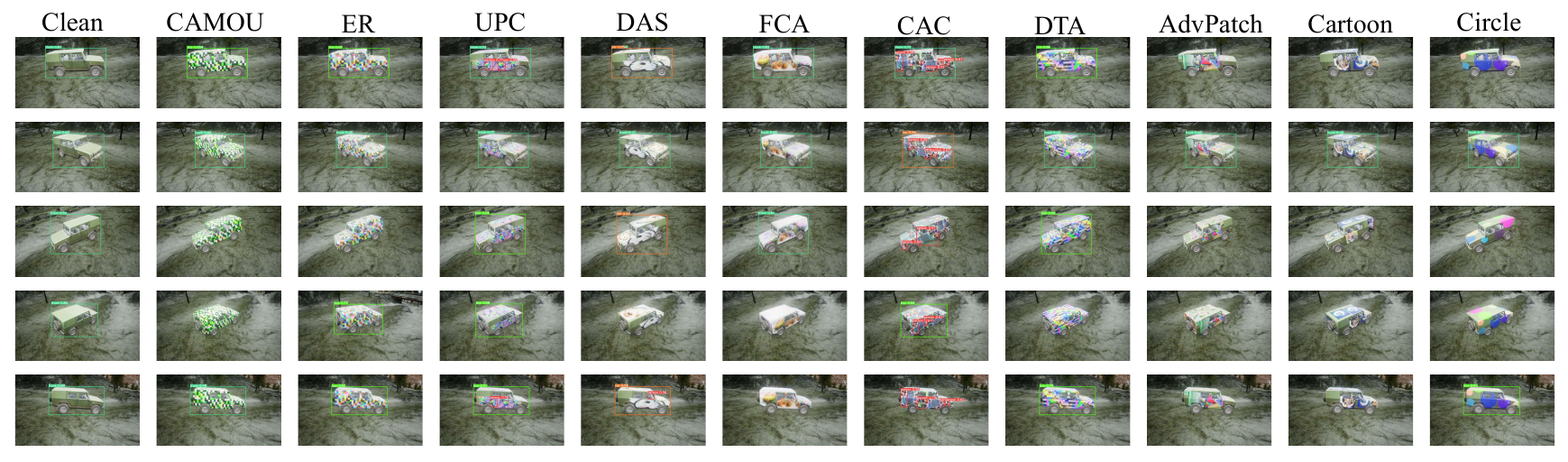}
	\end{minipage}
	\caption{Visualization of various adversarial textured vehicles (``truck") in a simulated environment. The detection results are given by YOLOv5x. Zoom in for details.}
	\label{fig:simluate_vehicle}
\end{figure*}
\subsection{Simulated adversarial attack}

In this part, we conduct the simulated adversarial attack experiment for vehicle object detection with the decided evaluated tool mentioned in Section \ref{sec:tool}. Specifically, we record a ten-second video with frame per second (FPS) 30 and extract 300 images for evaluation. Table \ref{tab:simluated_attack} lists the quantitative result of different adversarial textures for vehicle detection. As we can observe, the AdvPatch obtains superior attack performance except for CAC on the average decrease of P@0.5 by 30.83\% on four detectors. The possible reason is that CAC optimized the adversarial texture by exploiting the information of target models on a large-scale dataset. In contrast, the proposed black-box method is optimized on fewer images. On the other hand, the attack performance of our method (i.e., Circle and Cartoon) is on par with the existing approaches, indicating the proposed method's effectiveness. 

Additionally, the YOLOv5x shows vulnerable to adversarial texture attacks than other methods. Specifically, the average P@0.5 under various attack methods of YOLOv5x is 38.64\%, while Faster RCNN, Mask RCNN, and RetinaNet are 80.55\%, 84.42\%, and 63.26\%, respectively. Such findings indicate the robustness discrepancy across the detector, which can help people design more robust architecture. Figure \ref{fig:simluate_vehicle} illustrated the simulated adversarial examples.

\subsection{Ablation study}
In this part, we exhaustively investigate the influence of the following factors on the attack performance: important-aware selection strategy, the number of image pools, and the devised constraints composed of masking and overlap control.

\subsubsection{Important-aware selection strategy}
\begin{figure}[t]
	\centering
	\begin{minipage}{.9\linewidth}
		\centering
		\includegraphics[width =1\linewidth]{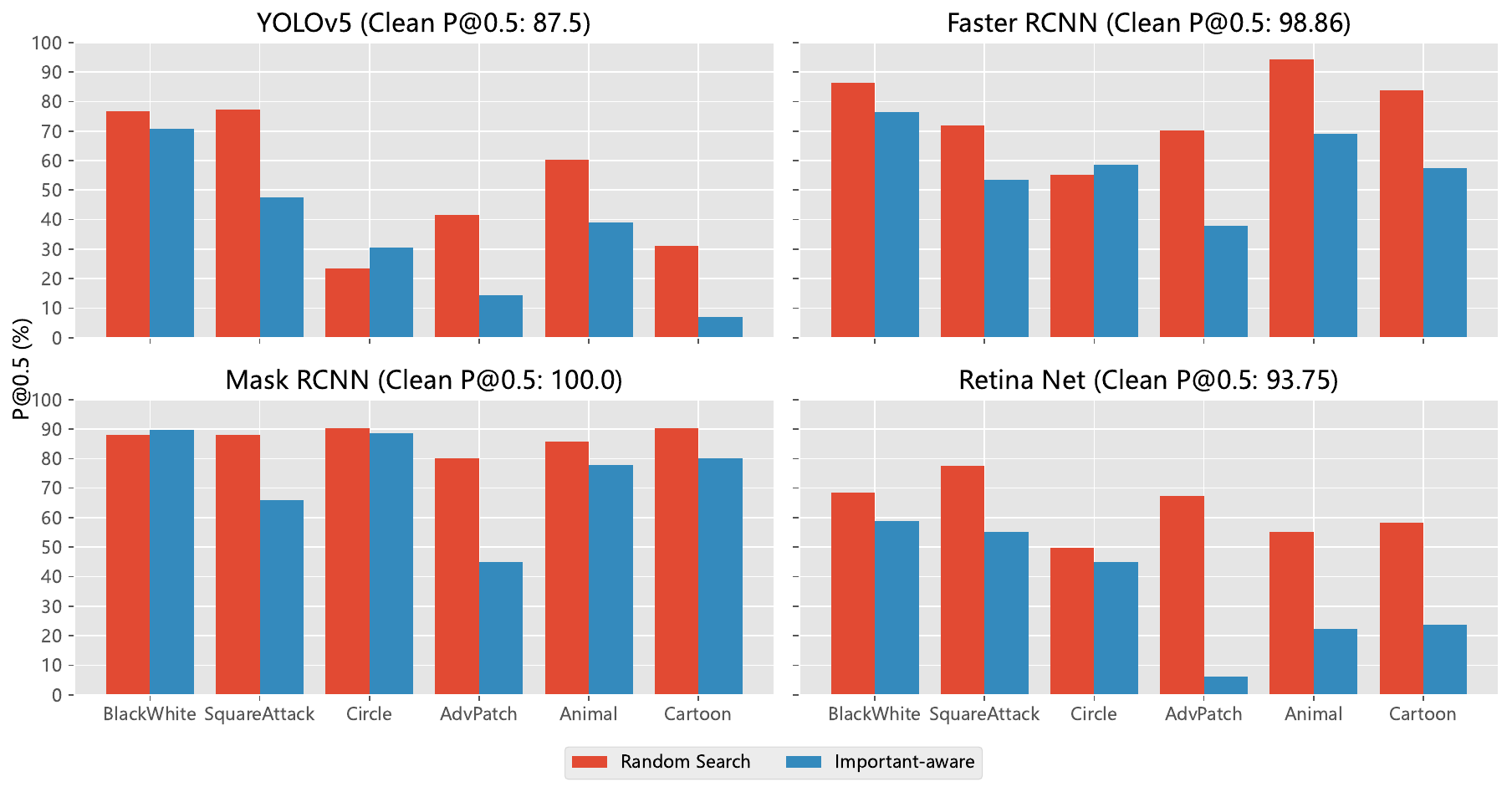}
	\end{minipage}
	\caption{Effectiveness of local search on attack performance.}
	\label{fig:ablation_local}
\end{figure}
\begin{figure}[!htbp]
	\centering
	\begin{minipage}{.9\linewidth}
		\centering
		\includegraphics[width =1\linewidth]{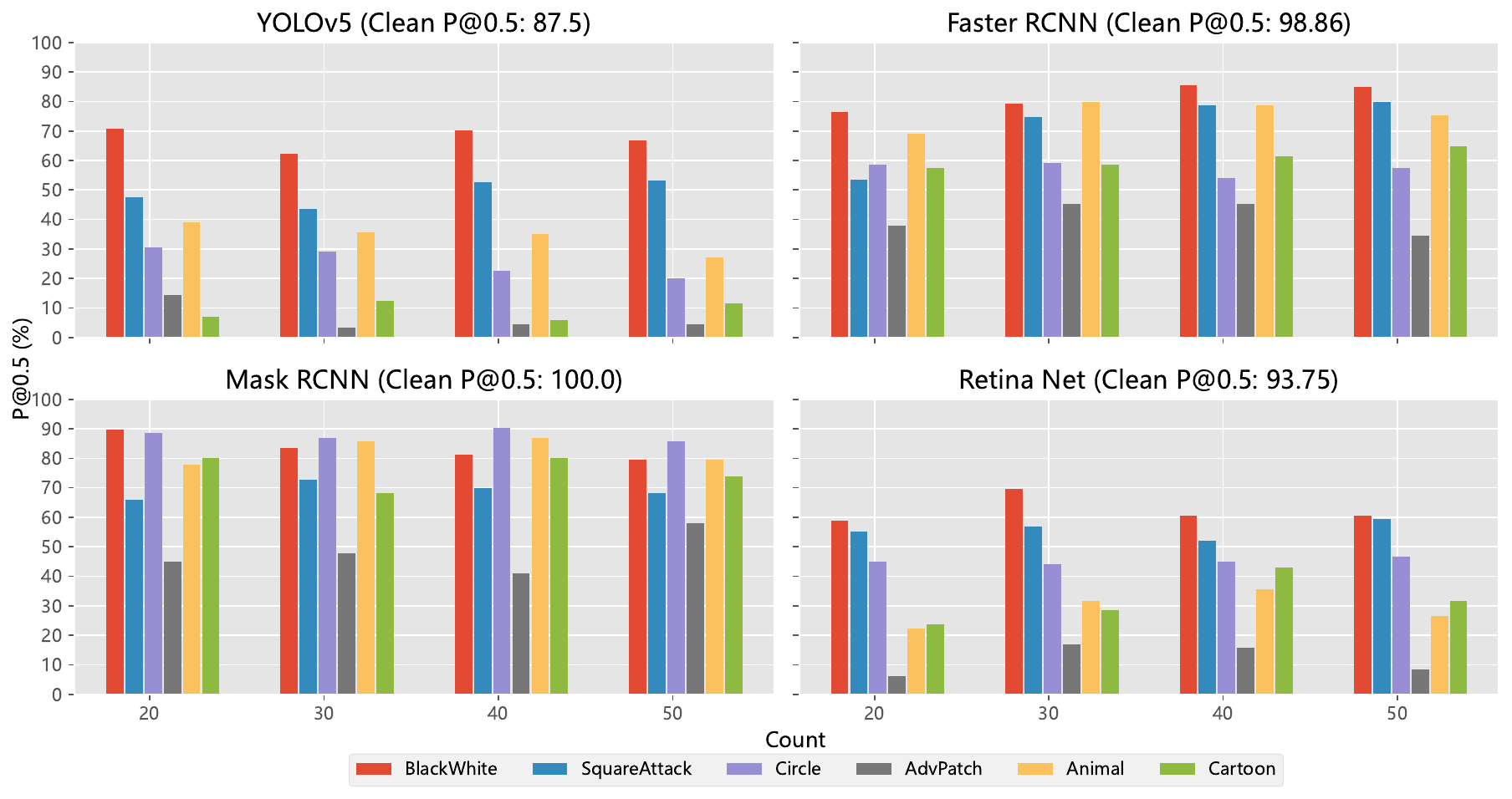}
	\end{minipage}
	\caption{Effectiveness of varying image count of image pools on attack performance. }
	\label{fig:ablation_image_count}
\end{figure}

The random search may fail to seek the best image patch for the specificity position, while the important-aware strategy is used to solve this problem. To study the effectiveness of the important-aware strategy, we perform experiments using the image pool of 20 images (i.e., $|S|=20$). Figure \ref{fig:ablation_local} depicts the comparison results. We can conclude that the local search can boost the attack performance of the proposed method. Specifically, we obtain the average gains of model performance degradation by 10\%. Meanwhile, we also observe that the important-aware strategy has minimum impact on BlackWhite, and Circle, which can be attributed to the pure color no content. However, the important-aware strategy obtains the attack improvement with the search time cost. Therefore, we can trade off both by switching to the important-aware strategy or not.

\subsubsection{Number of image pools}
Intuitively, enlarging the search space (i.e., diversity of the image pool) could benefit in obtaining a better solution. To verify this conjecture, we investigate the influence of different image numbers (i.e., {20, 30, 40, 50}) in the image pool on attack performance. Figure \ref{fig:ablation_image_count} describe the evaluation results. We can observe no obvious discrepancies when adopting the different number of images pool. For example,  the standard deviation of YOLOv5 on AdvPatch, Animal, and Cartoon on the different image numbers of image pools is 3.88, 3.76, and 2.46, respectively, which indicates that increasing the image number of the image pool can not bring significant improvement. Besides, we find an interesting phenomenon that the one-stage detectors are more fragile than the two-stage detectors under the proposed attack, which encourage people to pay more attention to the robustness of the one-stage detector.

\subsubsection{Effectiveness of the proposed constraint conditions}
\begin{figure}[t]
	\centering
	\begin{minipage}{.25\linewidth}
		\centering
		\includegraphics[width =1\linewidth]{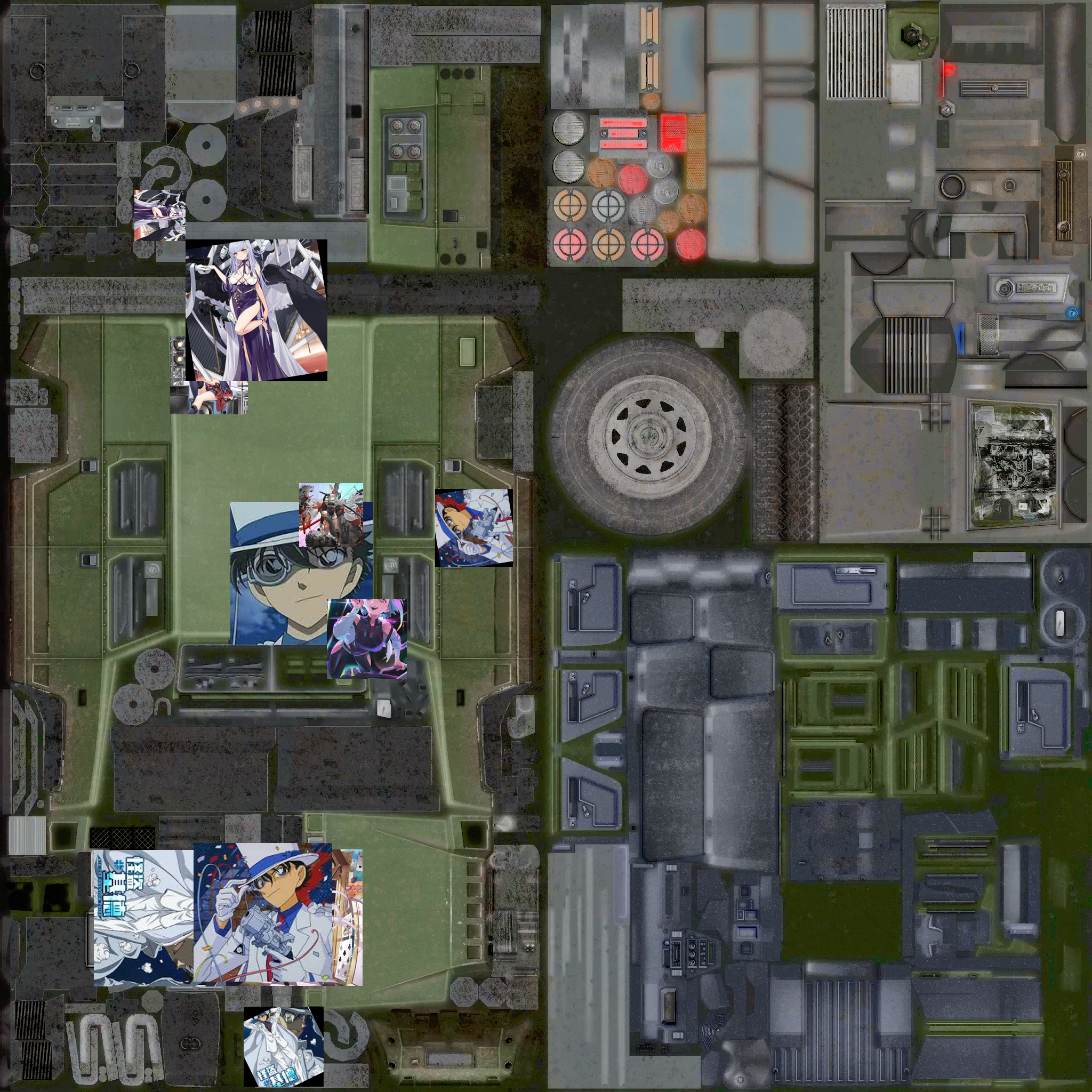}
		\centerline{\footnotesize w/o Overlap \& Mask}
	\end{minipage}
	\begin{minipage}{.25\linewidth}
		\centering
		\includegraphics[width =1\linewidth]{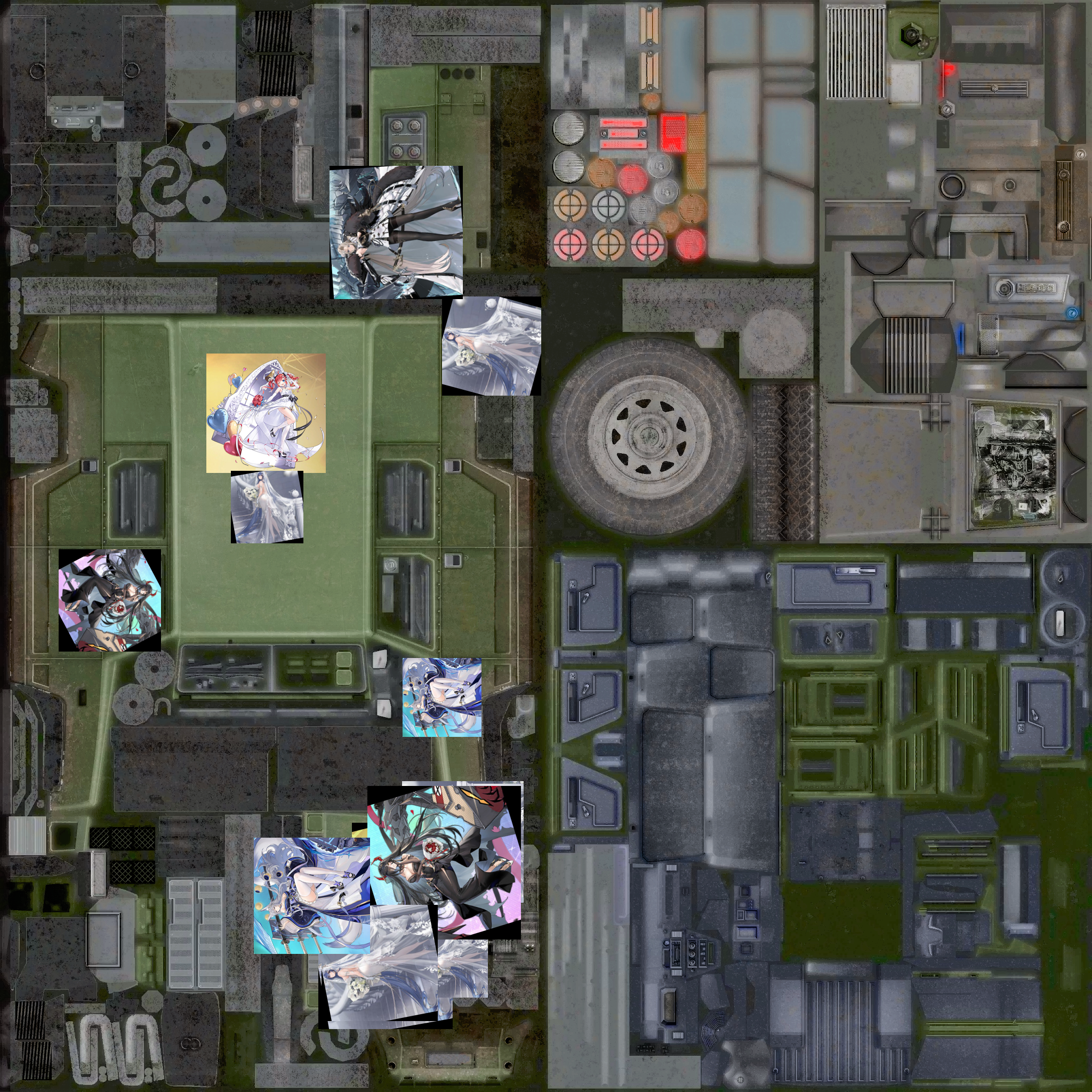}
		\centerline{\footnotesize w/o Overlap} 
	\end{minipage}
	\begin{minipage}{.25\linewidth}
		\centering
		\includegraphics[width =1\linewidth]{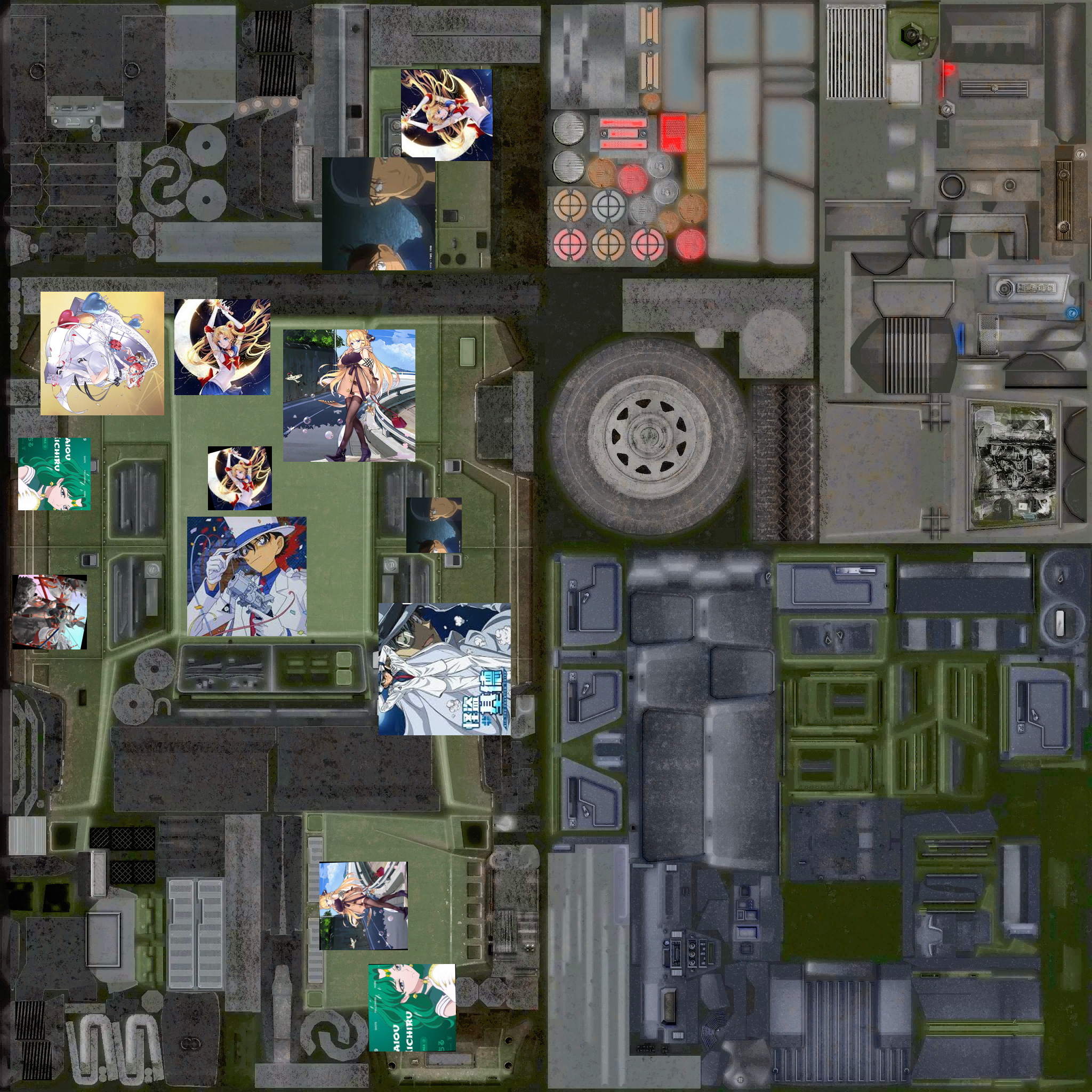}
		\centerline{\footnotesize Overlap \& Mask} 
	\end{minipage}

	\begin{minipage}{.5\linewidth}
		\centering
		\includegraphics[width =1\linewidth]{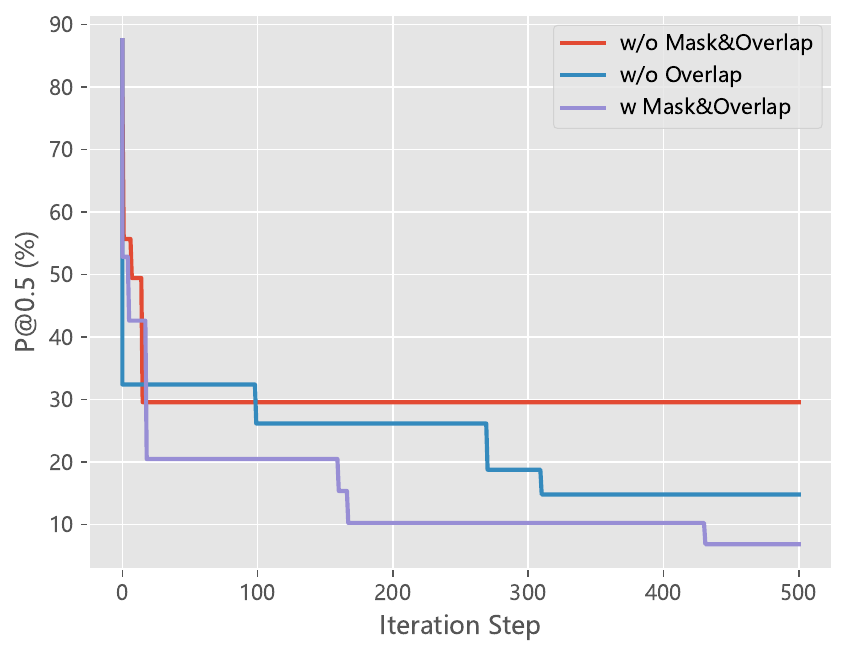}
	\end{minipage}
	\caption{Effectiveness of proposed variable constraints on attack performance.}
	\label{fig:ablation_constraints}
\end{figure}

The effectiveness and efficiency of the proposed method have been influenced by the search space, which is constrained by the proposed two constraint conditions. Thus, we optimize the adversarial texture against the YOLOv5 with the default experiment settings under the following constraints composed: without (w/o) Mask\&Overlap, w/o Overlap, and the complete method. Quantitative and qualitative results are illustrated in Figure \ref{fig:ablation_constraints}. As we can observe, on the one hand, the complete method not only convergences fast and obtains the best attack performance at the same iteration number. Specifically, the complete method degrades the detector's performance by 67.05\% at the 18-th iteration, while w/o Mask\&Overlap or w/o Overlap degrades by 57.95\% and 55.11\%, respectively. Moreover, the complete method deteriorates the detector's performance by 80.68\% at the 500-th iteration, while the other two are 57.95\% and 72.73\%. On the other hand, the w/o Overlap\&Mask place the image patch on the non-appearance region of texture, and w/o Overlap results in overlap phenomena. By contrast, the complete method obtains the rational layout of image patches.

\section{Extension of the proposed method}
In this section, we investigate the flexibility of the proposed method. Specifically, we perform two additional experiments, i.e., the warplane detection task and the traffic sign recognition task. The former is used to investigate whether the proposed method can be applied to find the optimal adversarial UV texture to the 3D model that has no explicit UV texture, while the latter is used to investigate whether the proposed method can be applied to find the optimal adversarial layout of image sticker on 2D images.

\subsection{Extend to warplane detection task}
To investigate the extensibility of the proposed method, we optimize the adversarial texture for no UV texture 3D model. Specifically, we first construct a base UV texture initialized with pure color. Then, we optimize the optimal layout of the elements on the base UV for adversarially. We discard the mask constraint as the 3D model has no UV texture. In this part, we collect 160 images sampled from different viewpoints as the test set, and the number of the image pool is 40. The evaluation results are listed in Table \ref{tab:comparison_noneuv}, as we can observe, on the one hand, the AdvPatch achieves the best result against all models (i.e., YOLOv5x, Faster RCNN, Mask RCNN, and RetinaNet), obtaining the maximum drop in terms of P@0.5 by 83.75\%, 82.5\%, 78.75\%, and 82.5\%, respectively. On the other hand, the average decrease in terms of P@0.5 of four detectors exceeds 50\%, i.e., 54.53\% and 54.69\% for Circle and Cartoon. The experiment result shows the effectiveness of the proposed method in generating the adversarial texture for the none-UV texture 3D model.

\begin{table}[t]
\centering
\scriptsize
\setlength\tabcolsep{4pt}
\caption{Evaluation results of various \textit{digital attacks} in terms of P@0.5 (\%) for \textit{warplane}. \textbf{Bold} item highlights the best result, where the item in the bracket denotes the gain of P@0.5.}
\label{tab:comparison_noneuv}
\begin{tabular}{ccccc}
\hline
              & YOLOv5x & FasterRCNN & MaskRCNN & RetinaNet \\ \hline
RAW           & 95.00      & 91.25      & 88.75    & 86.88     \\
BlackWhite \cite{eykholt2018robust}    & 81.25($\downarrow$13.75)   & 57.50 ($\downarrow$33.75)     & 50.63($\downarrow$38.12)    & 39.38($\downarrow$47.50)     \\
SquareAttack \cite{andriushchenko2020square}  & 48.75($\downarrow$46.25)   & 24.38($\downarrow$66.87)      & 31.25($\downarrow$57.5)     & 33.13($\downarrow$53.75)     \\
Circle        & 40.63($\downarrow$54.37)   & 31.88($\downarrow$59.37)      & 31.88($\downarrow$56.87)    & 39.38($\downarrow$47.50)     \\
AdvPatch      & 11.25($\downarrow$\textbf{83.75}) & ~8.75 ($\downarrow$\textbf{82.50}) & 10.00   ($\downarrow$\textbf{78.75})    & ~4.38 ($\downarrow$\textbf{82.50})     \\
Animal        & 61.88($\downarrow$33.12)   & 49.38($\downarrow$41.87)      & 50.63($\downarrow$38.12)    & 48.75($\downarrow$38.13)     \\
Cartoon       & 33.13($\downarrow$61.87)   & 40.63($\downarrow$50.62)      & 46.88($\downarrow$41.87)    & 22.50($\downarrow$64.38)     \\ \hline
\end{tabular}
\end{table}

\begin{table}[t]
\centering
\scriptsize
\setlength\tabcolsep{4pt}
\caption{Evaluation results of various \textit{simulated attacks} in terms of P@0.5 (\%) for \textit{warplane}. \textbf{Bold} item highlights the best result, where the item in the bracket denotes the gain of P@0.5.}
\label{tab:simluated_warplane}
\begin{tabular}{ccccc}
\hline
              & YOLOv5x & FasterRCNN & MaskRCNN & RetinaNet \\ \hline
RAW           & 100.0      & 100.0      & 100.0    & 100.0     \\
BlackWhite \cite{eykholt2018robust}    & 98.34($\downarrow$1.66)    & 99.00($\downarrow$33.75)      & 99.67($\downarrow$~0.33)    & 99.67($\downarrow$~0.33)     \\
SquareAttack \cite{andriushchenko2020square}  & 58.80($\downarrow$41.20)   & 22.26($\downarrow$77.74)      & 69.10($\downarrow$30.90)    & 82.06($\downarrow$17.94)     \\
Circle        & 61.13($\downarrow$38.87)   & 32.56($\downarrow$67.44)      & 79.40($\downarrow$20.60)    & 83.06($\downarrow$16.94)     \\
AdvPatch      & 11.96($\downarrow$\textbf{88.04})   & ~9.63 ($\downarrow$\textbf{90.37})       & ~4.32  ($\downarrow$\textbf{95.68})    & 12.96 ($\downarrow$\textbf{87.04})     \\
Animal        & 97.34($\downarrow$~2.66)   & 99.67($\downarrow$~0.33)      & 100.0($\downarrow$38.12)    & 99.00($\downarrow$~1.00)     \\
Cartoon       & 76.08($\downarrow$23.92)   & 86.71($\downarrow$13.29)      & 88.04($\downarrow$11.96)    & 96.01($\downarrow$~3.99)     \\ \hline
\end{tabular}
\end{table}

\begin{figure}[t]
	\centering
	\begin{minipage}{.9\linewidth}
		\centering
		\includegraphics[width =1\linewidth]{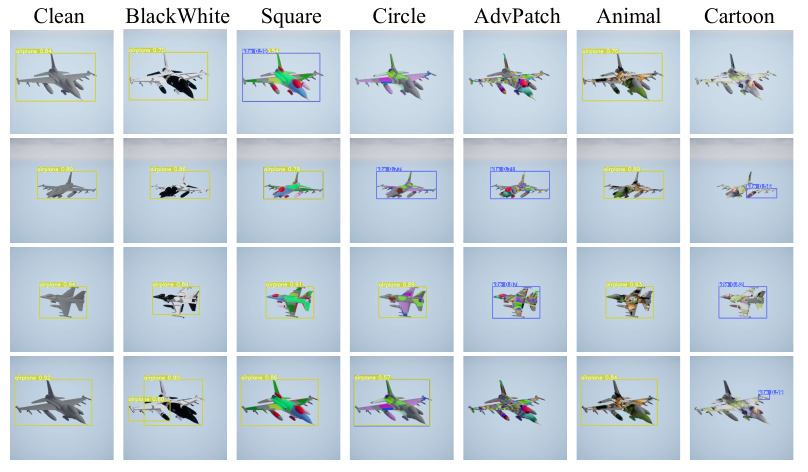}
	\end{minipage}
	\caption{Examples generated by different adversarial textures in a simulated environment against YOLOv5x.}
	\label{fig:simulate_warplane}
\end{figure}

Furthermore, we also conduct the simulated adversarial attack. Specifically, we use the evaluated tool to collect 301 images for evaluation. The evaluation results are illustrated in Table \ref{tab:simluated_warplane}. As we can see, AdvPatch achieves the best attack performance on four detectors, where the average decrease of P@0.5 is 90.28\%. Moreover, the adversarial texture optimized on the pure color displays superior attack performance (i.e., the average drop of P@0.5 is 16.53\% and 19.52\% for Square and Circle) than Animal (0.42\%) or Cartoon (13.29\%). The possible reason is that the warplane model has no UV texture, resulting in severe distortion of the natural image during pasting on the customed texture. In contrast, the adversarial texture optimized by the pure color exhibits better attack performance due to its small image distortion. Additionally, Figure \ref{fig:simulate_warplane} provides some simulated examples.

\subsection{Extend to traffic sign recognition}

\begin{table}[t]
\centering
\setlength\tabcolsep{2pt}
\caption{Evaluation results of our method on traffic sign recognition in terms of model's ASR.}
\label{tab:attack_tsr}
\begin{tabular}{ccccccc}
\hline
			 & BlackWhite & SquareAttack & Circle & AdvPatch & Animal & Cartoon  \\ \hline
GTSRB CNN   & 97\%  & 98\% & 90\% & 54\% & 76\% & 99\% \\ 
LISA CNN    & 98\% & 94\% & 80\% & 64\% & 90\% & 69\% \\ \hline
\end{tabular}
\end{table}
We extend our method to attack the traffic sign recognition (TSR) model to demonstrate not only the proposed method can seek the optimal adversarial layout of UV texture for the 3D objects but is also applicable to 2D images. Specifically, we first train two widely used TSR models consisting of LISA CNN and GTSRB CNN. Note that, we only focus on the ``STOP" sign category. Our goal is to optimize the universal adversarial image patch layout for the ``STOP" sign to mislead the TSR model to output the wrong result, namely the untargeted attack. We select 100 visually recognizable ``STOP" sign images from the test dataset for evaluation. 

The evaluation results are reported in Table \ref{tab:attack_tsr} regarding attack success rate (ASR). As we can see, the pure color baselines outperform the image-based methods, where the former achieves the average ASR of 94.5\%, and the latter is 75.33\%. We speculate the discrepancy between the TSR model and the vehicle detection model in model architecture and task specificity leads to the gap in attack performance. Moreover, the pure color baselines have bigger search space than the image-based method, which is effective when attacking the TSR model. Additionally, we also observe that the AdvPatch performs worse than other methods, which can be attributed to the AdvPatch being originally designed to attack object detectors, while TSR is a classification task.

\section{Conclusion}
\label{sec:conclusion}
In this paper, we proposed a novel universal multi-view black-box attack against the object detector. Specifically, we instead optimize the adversarial layout constructed by multiple stickers for a 3D object's texture, whose rendered images can deceive the object detector from the multi-view scenario. Moreover, the image sticker can be arbitrary images, which renders our attack more inconspicuous. To find the adversarial layout, we first model the above problem as a circle-based layout optimization problem and then exploit the random search algorithm enhanced by the proposed important-aware selection strategy to find the optimal sticker layout. Additionally, we proposed a photo-realistic simulator-based evaluation tool to assess the UV texture-based attack approaches. Extensive digital and simulated experiments, as well as extended experiments, suggested the effectiveness and extensibility of the proposed method.

\bibliographystyle{IEEEtran}
\bibliography{references}

\begin{thebibliography}{10}
\providecommand{\url}[1]{#1}
\csname url@samestyle\endcsname
\providecommand{\newblock}{\relax}
\providecommand{\bibinfo}[2]{#2}
\providecommand{\BIBentrySTDinterwordspacing}{\spaceskip=0pt\relax}
\providecommand{\BIBentryALTinterwordstretchfactor}{4}
\providecommand{\BIBentryALTinterwordspacing}{\spaceskip=\fontdimen2\font plus
\BIBentryALTinterwordstretchfactor\fontdimen3\font minus
  \fontdimen4\font\relax}
\providecommand{\BIBforeignlanguage}[2]{{%
\expandafter\ifx\csname l@#1\endcsname\relax
\typeout{** WARNING: IEEEtran.bst: No hyphenation pattern has been}%
\typeout{** loaded for the language `#1'. Using the pattern for}%
\typeout{** the default language instead.}%
\else
\language=\csname l@#1\endcsname
\fi
#2}}
\providecommand{\BIBdecl}{\relax}
\BIBdecl

\bibitem{szegedy2014intriguing}
\BIBentryALTinterwordspacing
C.~Szegedy, W.~Zaremba, I.~Sutskever, J.~Bruna, D.~Erhan, I.~J. Goodfellow, and
  R.~Fergus, ``Intriguing properties of neural networks,'' in \emph{2nd
  International Conference on Learning Representations, {ICLR} 2014, Banff, AB,
  Canada, April 14-16, 2014, Conference Track Proceedings}, 2014. [Online].
  Available: \url{http://arxiv.org/abs/1312.6199}
\BIBentrySTDinterwordspacing

\bibitem{pgd2018towards}
A.~Madry, A.~Makelov, L.~Schmidt, D.~Tsipras, and A.~Vladu, ``Towards deep
  learning models resistant to adversarial attacks,'' in \emph{6th
  International Conference on Learning Representations, {ICLR} 2018, Vancouver,
  BC, Canada, April 30 - May 3, 2018, Conference Track Proceedings}.\hskip 1em
  plus 0.5em minus 0.4em\relax OpenReview.net, 2018.

\bibitem{mim2018BoostingAA}
Y.~Dong, F.~Liao, T.~Pang, H.~Su, J.~Zhu, X.~Hu, and J.~Li, ``Boosting
  adversarial attacks with momentum,'' \emph{2018 IEEE/CVF Conference on
  Computer Vision and Pattern Recognition}, pp. 9185--9193, 2018.

\bibitem{li2022approximate}
C.~Li, H.~Wang, J.~Zhang, W.~Yao, and T.~Jiang, ``An approximated gradient sign
  method using differential evolution for black-box adversarial attack,''
  \emph{IEEE Transactions on Evolutionary Computation}, pp. 1--1, 2022.

\bibitem{li2023adaptive}
C.~Li, W.~Yao, H.~Wang, and T.~Jiang, ``Adaptive momentum variance for
  attention-guided sparse adversarial attacks,'' \emph{Pattern Recognition},
  vol. 133, p. 108979, 2023.

\bibitem{wang2023rfla}
D.~Wang, W.~Yao, T.~Jiang, C.~Li, and X.~Chen, ``Rfla: A stealthy reflected
  light adversarial attack in the physical world,'' in \emph{Proceedings of the
  IEEE/CVF International Conference on Computer Vision}, 2023, pp. 4455--4465.

\bibitem{fgsm2015explaining}
\BIBentryALTinterwordspacing
I.~Goodfellow, J.~Shlens, and C.~Szegedy, ``Explaining and harnessing
  adversarial examples,'' in \emph{International Conference on Learning
  Representations}, 2015. [Online]. Available:
  \url{http://arxiv.org/abs/1412.6572}
\BIBentrySTDinterwordspacing

\bibitem{bim2016adversarial}
A.~Kurakin, I.~Goodfellow, and S.~Bengio, ``Adversarial machine learning at
  scale,'' \emph{arXiv preprint arXiv:1611.01236}, 2016.

\bibitem{tang2023natural}
G.~Tang, W.~Yao, T.~Jiang, W.~Zhou, Y.~Yang, and D.~Wang, ``Natural
  weather-style black-box adversarial attacks against optical aerial
  detectors,'' \emph{IEEE Transactions on Geoscience and Remote Sensing},
  vol.~61, pp. 1--11, 2023.

\bibitem{thys2019fooling}
S.~Thys, W.~Van~Ranst, and T.~Goedem{\'e}, ``Fooling automated surveillance
  cameras: adversarial patches to attack person detection,'' in
  \emph{Proceedings of the IEEE/CVF conference on computer vision and pattern
  recognition workshops}, 2019, pp. 0--0.

\bibitem{huang2020universal}
L.~Huang, C.~Gao, Y.~Zhou, C.~Xie, A.~L. Yuille, C.~Zou, and N.~Liu,
  ``Universal physical camouflage attacks on object detectors,'' in
  \emph{Proceedings of the IEEE/CVF Conference on Computer Vision and Pattern
  Recognition}, 2020, pp. 720--729.

\bibitem{wang2021dual}
J.~Wang, A.~Liu, Z.~Yin, S.~Liu, S.~Tang, and X.~Liu, ``Dual attention
  suppression attack: Generate adversarial camouflage in physical world,'' in
  \emph{Proceedings of the IEEE/CVF Conference on Computer Vision and Pattern
  Recognition}, 2021, pp. 8565--8574.

\bibitem{wang2022fca}
D.~Wang, T.~Jiang, J.~Sun, W.~Zhou, Z.~Gong, X.~Zhang, W.~Yao, and X.~Chen,
  ``Fca: Learning a 3d full-coverage vehicle camouflage for multi-view physical
  adversarial attack,'' in \emph{Proceedings of the AAAI Conference on
  Artificial Intelligence}, vol.~36, no.~2, 2022, pp. 2414--2422.

\bibitem{duan2022learning}
Y.~Duan, J.~Chen, X.~Zhou, J.~Zou, Z.~He, J.~Zhang, W.~Zhang, and Z.~Pan,
  ``Learning coated adversarial camouflages for object detectors,'' in
  \emph{Proceedings of the Thirty-First International Joint Conference on
  Artificial Intelligence, {IJCAI} 2022, Vienna, Austria, 23-29 July 2022},
  L.~D. Raedt, Ed., 2022, pp. 891--897.

\bibitem{suryanto2022dta}
N.~Suryanto, Y.~Kim, H.~Kang, H.~T. Larasati, Y.~Yun, T.-T.-H. Le, H.~Yang,
  S.-Y. Oh, and H.~Kim, ``Dta: Physical camouflage attacks using differentiable
  transformation network,'' in \emph{Proceedings of the IEEE/CVF Conference on
  Computer Vision and Pattern Recognition}, 2022, pp. 15\,305--15\,314.

\bibitem{wu2020physical}
T.~Wu, X.~Ning, W.~Li, R.~Huang, H.~Yang, and Y.~Wang, ``Physical adversarial
  attack on vehicle detector in the carla simulator,'' \emph{arXiv preprint
  arXiv:2007.16118}, 2020.

\bibitem{liu2019perceptual}
A.~Liu, X.~Liu, J.~Fan, Y.~Ma, A.~Zhang, H.~Xie, and D.~Tao,
  ``Perceptual-sensitive gan for generating adversarial patches,'' in
  \emph{Proceedings of the AAAI conference on artificial intelligence},
  vol.~33, no.~01, 2019, pp. 1028--1035.

\bibitem{doan2022tnt}
B.~G. Doan, M.~Xue, S.~Ma, E.~Abbasnejad, and D.~C. Ranasinghe, ``Tnt attacks!
  universal naturalistic adversarial patches against deep neural network
  systems,'' \emph{IEEE Transactions on Information Forensics and Security},
  2022.

\bibitem{wei2022adversarial}
X.~Wei, Y.~Guo, and J.~Yu, ``Adversarial sticker: A stealthy attack method in
  the physical world,'' \emph{IEEE Transactions on Pattern Analysis and Machine
  Intelligence}, 2022.

\bibitem{chen2018shapeshifter}
S.-T. Chen, C.~Cornelius, J.~Martin, and D.~H.~P. Chau, ``Shapeshifter: Robust
  physical adversarial attack on faster r-cnn object detector,'' in \emph{Joint
  European Conference on Machine Learning and Knowledge Discovery in
  Databases}.\hskip 1em plus 0.5em minus 0.4em\relax Springer, 2018, pp.
  52--68.

\bibitem{xie2021improving}
P.~Xie, L.~Wang, R.~Qin, K.~Qiao, S.~Shi, G.~Hu, and B.~Yan, ``Improving the
  transferability of adversarial examples with new iteration framework and
  input dropout,'' \emph{arXiv preprint arXiv:2106.01617}, 2021.

\bibitem{alzantot2019genattack}
M.~Alzantot, Y.~Sharma, S.~Chakraborty, H.~Zhang, C.-J. Hsieh, and M.~B.
  Srivastava, ``Genattack: Practical black-box attacks with gradient-free
  optimization,'' in \emph{Proceedings of the Genetic and Evolutionary
  Computation Conference}, 2019, pp. 1111--1119.

\bibitem{lin2020black}
J.~Lin, L.~Xu, Y.~Liu, and X.~Zhang, ``Black-box adversarial sample generation
  based on differential evolution,'' \emph{Journal of Systems and Software},
  vol. 170, p. 110767, 2020.

\bibitem{zhang2019attacking}
Q.~Zhang, K.~Wang, W.~Zhang, and J.~Hu, ``Attacking black-box image classifiers
  with particle swarm optimization,'' \emph{IEEE Access}, vol.~7, pp.
  158\,051--158\,063, 2019.

\bibitem{yang2020patchattack}
C.~Yang, A.~Kortylewski, C.~Xie, Y.~Cao, and A.~Yuille, ``Patchattack: A
  black-box texture-based attack with reinforcement learning,'' in
  \emph{European Conference on Computer Vision}.\hskip 1em plus 0.5em minus
  0.4em\relax Springer, 2020, pp. 681--698.

\bibitem{andriushchenko2020square}
M.~Andriushchenko, F.~Croce, N.~Flammarion, and M.~Hein, ``Square attack: a
  query-efficient black-box adversarial attack via random search,'' in
  \emph{European Conference on Computer Vision}.\hskip 1em plus 0.5em minus
  0.4em\relax Springer, 2020, pp. 484--501.

\bibitem{zhou2016learning}
B.~Zhou, A.~Khosla, A.~Lapedriza, A.~Oliva, and A.~Torralba, ``Learning deep
  features for discriminative localization,'' in \emph{Proceedings of the IEEE
  conference on computer vision and pattern recognition}, 2016, pp. 2921--2929.

\bibitem{moosavi2017universal}
S.-M. Moosavi-Dezfooli, A.~Fawzi, O.~Fawzi, and P.~Frossard, ``Universal
  adversarial perturbations,'' in \emph{Proceedings of the IEEE conference on
  computer vision and pattern recognition}, 2017, pp. 1765--1773.

\bibitem{mopuri2017fast}
K.~R. Mopuri, U.~Garg, and R.~V. Babu, ``Fast feature fool: A data independent
  approach to universal adversarial perturbations,'' in \emph{Proceedings of
  the British Machine Vision Conference ({BMVC})}, 2017.

\bibitem{mopuri2018generalizable}
K.~R. Mopuri, A.~Ganeshan, and R.~V. Babu, ``Generalizable data-free objective
  for crafting universal adversarial perturbations,'' \emph{IEEE transactions
  on pattern analysis and machine intelligence}, vol.~41, no.~10, pp.
  2452--2465, 2018.

\bibitem{mopuri2018ask}
K.~R. Mopuri, P.~K. Uppala, and R.~V. Babu, ``Ask, acquire, and attack:
  Data-free uap generation using class impressions,'' in \emph{Proceedings of
  the European Conference on Computer Vision (ECCV)}, 2018, pp. 19--34.

\bibitem{wang2023improving}
D.~Wang, W.~Yao, T.~Jiang, and X.~Chen, ``Improving transferability of
  universal adversarial perturbation with feature disruption,'' \emph{IEEE
  Transactions on Image Processing}, 2023.

\bibitem{dong2022viewfool}
Y.~Dong, S.~Ruan, H.~Su, C.~Kang, X.~Wei, and J.~Zhu, ``Viewfool: Evaluating
  the robustness of visual recognition to adversarial viewpoints,'' in
  \emph{Advances in Neural Information Processing Systems}, A.~H. Oh,
  A.~Agarwal, D.~Belgrave, and K.~Cho, Eds., 2022.

\bibitem{ghosh2022black}
A.~Ghosh, S.~S. Mullick, S.~Datta, S.~Das, A.~K. Das, and R.~Mallipeddi, ``A
  black-box adversarial attack strategy with adjustable sparsity and
  generalizability for deep image classifiers,'' \emph{Pattern Recognition},
  vol. 122, p. 108279, 2022.

\bibitem{wang2022survey}
D.~Wang, W.~Yao, T.~Jiang, G.~Tang, and X.~Chen, ``A survey on physical
  adversarial attack in computer vision,'' \emph{arXiv preprint
  arXiv:2209.14262}, 2022.

\bibitem{tan2021legitimate}
J.~Tan, N.~Ji, H.~Xie, and X.~Xiang, ``Legitimate adversarial patches: Evading
  human eyes and detection models in the physical world,'' in \emph{Proceedings
  of the 29th ACM International Conference on Multimedia}, 2021, pp.
  5307--5315.

\bibitem{hu2021naturalistic}
Y.-C.-T. Hu, B.-H. Kung, D.~S. Tan, J.-C. Chen, K.-L. Hua, and W.-H. Cheng,
  ``Naturalistic physical adversarial patch for object detectors,'' in
  \emph{Proceedings of the IEEE/CVF International Conference on Computer
  Vision}, 2021, pp. 7848--7857.

\bibitem{zhang2019camou}
Y.~Zhang, P.~H. Foroosh, and B.~Gong, ``Camou: Learning a vehicle camouflage
  for physical adversarial attack on object detections in the wild,''
  \emph{ICLR}, 2019.

\bibitem{glenn_jocher_2021_5563715}
\BIBentryALTinterwordspacing
G.~J. et. al., ``{ultralytics/yolov5: v6.0 - YOLOv5n 'Nano' models, Roboflow
  integration, TensorFlow export, OpenCV DNN support},'' 2021. [Online].
  Available: \url{https://doi.org/10.5281/zenodo.5563715}
\BIBentrySTDinterwordspacing

\bibitem{lin2017focal}
T.-Y. Lin, P.~Goyal, R.~Girshick, K.~He, and P.~Doll{\'a}r, ``Focal loss for
  dense object detection,'' in \emph{Proceedings of the IEEE international
  conference on computer vision}, 2017, pp. 2980--2988.

\bibitem{ren2015faster}
S.~Ren, K.~He, R.~Girshick, and J.~Sun, ``Faster r-cnn: Towards real-time
  object detection with region proposal networks,'' \emph{Advances in neural
  information processing systems}, vol.~28, pp. 91--99, 2015.

\bibitem{he2017mask}
K.~He, G.~Gkioxari, P.~Doll{\'a}r, and R.~Girshick, ``Mask r-cnn,'' in
  \emph{Proceedings of the IEEE international conference on computer vision},
  2017, pp. 2961--2969.

\bibitem{paszke2019pytorch}
A.~Paszke, S.~Gross, F.~Massa, A.~Lerer, J.~Bradbury, G.~Chanan, T.~Killeen,
  Z.~Lin, N.~Gimelshein, L.~Antiga \emph{et~al.}, ``Pytorch: An imperative
  style, high-performance deep learning library,'' \emph{Advances in neural
  information processing systems}, vol.~32, 2019.

\bibitem{lin2014microsoft}
T.-Y. Lin, M.~Maire, S.~Belongie, J.~Hays, P.~Perona, D.~Ramanan,
  P.~Doll{\'a}r, and C.~L. Zitnick, ``Microsoft coco: Common objects in
  context,'' in \emph{European conference on computer vision}.\hskip 1em plus
  0.5em minus 0.4em\relax Springer, 2014, pp. 740--755.

\bibitem{gtsrb2011}
J.~Stallkamp, M.~Schlipsing, J.~Salmen, and C.~Igel, ``The german traffic sign
  recognition benchmark: a multi-class classification competition,'' in
  \emph{The 2011 international joint conference on neural networks}.\hskip 1em
  plus 0.5em minus 0.4em\relax IEEE, 2011, pp. 1453--1460.

\bibitem{lisa2012}
A.~Mogelmose, M.~M. Trivedi, and T.~B. Moeslund, ``Vision-based traffic sign
  detection and analysis for intelligent driver assistance systems:
  Perspectives and survey,'' \emph{IEEE Transactions on Intelligent
  Transportation Systems}, vol.~13, no.~4, pp. 1484--1497, 2012.

\bibitem{Zhou2018}
Q.-Y. Zhou, J.~Park, and V.~Koltun, ``Open3d: A modern library for 3d data
  processing,'' \emph{arXiv:1801.09847}, 2018.

\bibitem{eykholt2018robust}
K.~Eykholt, I.~Evtimov, E.~Fernandes, B.~Li, A.~Rahmati, C.~Xiao, A.~Prakash,
  T.~Kohno, and D.~Song, ``Robust physical-world attacks on deep learning
  visual classification,'' in \emph{Proceedings of the IEEE conference on
  computer vision and pattern recognition}, 2018, pp. 1625--1634.

\end{thebibliography}

\end{document}